\documentclass[acmsmall]{acmart}

\AtBeginDocument{%
  }

\setcopyright{cc}
\setcctype{by}
\acmDOI{10.1145/3832146}
\acmYear{2026}
\acmJournal{PACMSE}
\acmVolume{3}
\acmNumber{ISSTA}
\acmArticle{ISSTA055}
\acmMonth{10}
\acmSubmissionID{issta26main-p456-p}
\received{2026-01-30}
\received[accepted]{2026-06-25}

\usepackage{latexsym}
\usepackage[T1]{fontenc}
\usepackage[utf8]{inputenc}
\usepackage{microtype}
\usepackage{graphicx}
\usepackage{pifont}
\usepackage{xspace}
\usepackage{amsmath} 
\usepackage{amsfonts}
\usepackage{amsthm}
\usepackage{xcolor}
\usepackage[dvipsnames]{xcolor}
\usepackage{fontawesome5}
\usepackage[most]{tcolorbox}
\tcbuselibrary{breakable}
\usepackage{wrapfig}
\usepackage{graphicx}
\usepackage[percent]{overpic}
\usepackage{enumitem}
\usepackage{subcaption}
\usepackage{multirow} 
\usepackage[table]{xcolor}
\usepackage{algorithm}       
\usepackage{algorithmicx}     
\usepackage{algpseudocode}  
\usepackage{wrapfig}
\usepackage{makecell}
\usepackage{colortbl}     
\usepackage{booktabs}   

\def\name{\textit{\textsc{LAVE}}\xspace}
\newcommand{\baselinename}[1]{\textsc{#1}\xspace}
\newcommand{\llmname}[1]{{\fontfamily{pcr}\selectfont {#1}}\xspace}
\newcommand{\dataname}[1]{{\fontfamily{cmtt}\selectfont {#1}}\xspace}

\newif\ifdraftmode
\draftmodefalse

\usepackage[most]{tcolorbox}
\usepackage{xcolor}

\newtcbox{\codebox}{on line,
  colback=gray!10, colframe=gray!60,
  boxsep=1pt, left=2pt, right=2pt, top=1pt, bottom=1pt,
  fontupper=\ttfamily\small, arc=3pt}

\renewcommand{\cite}{\citep}

\def\eg{\emph{e.g.,}\xspace} 

\def\ie{\emph{i.e.,}\xspace}

\newtcolorbox{boxK}{
    top=2.4pt,
    bottom=2.4pt,
    left=4.5pt,
    right=4.5pt,
    boxrule = 0pt,
    toprule = 0pt, 
    enhanced,
    fuzzy shadow = {0pt}{-1pt}{-0.2pt}{0.2pt}{black!35} 
}

\newcommand{\cl}[1]{\textcolor{black}{#1}} 
\newenvironment{cp}[0]{\par\color{black}}{\par} 
\newcommand{\cc}[0]{\color{black}}

\newcommand{\up}[1]{\textsuperscript{\textcolor[HTML]{8B0000}{#1}}}
\newcommand{\same}[1]{\textsuperscript{\textcolor[HTML]{909090}{#1}}}
\newcommand{\down}[1]{\textsuperscript{\textcolor[HTML]{00009B}{#1}}}

\begin{document}









\title{Lookahead-Then-Verify: Reliable Constrained Decoding for Diffusion LLMs under Context-Free Grammars}

\author{Yitong Zhang}
\orcid{0009-0000-1138-4503}
\affiliation{%
  \institution{Tsinghua University}
  \department{College of AI}
  \city{Beijing}
  \country{China}
}
\email{22373337@buaa.edu.cn}

\author{Yongmin Li}
\orcid{0009-0001-3702-0043}
\affiliation{%
  \institution{Peking University}
  \department{School of Computer Science}
  \city{Beijing}
  \country{China}
}
\email{liyongmin@pku.edu.cn}

\author{Yuetong Liu}
\orcid{0009-0003-6512-2412}
\affiliation{%
  \institution{Tsinghua University}
  \department{College of AI}
  \city{Beijing}
  \country{China}
}
\email{23371522@buaa.edu.cn}

\author{Jia Li}
\orcid{0000-0002-5579-8852}
\affiliation{%
  \institution{Tsinghua University}
  \department{College of AI}
  \city{Beijing}
  \country{China}
}
\email{jia\_li@mail.tsinghua.edu.cn}

\author{Xiaoran Jia}
\orcid{0009-0005-8133-2502}
\affiliation{%
  \institution{Beijing Institute of Technology}
  \department{School of Computer Science and Technology}
  \city{Beijing}
  \country{China}
}
\email{jiaxiaoran576@gmail.com}

\author{Zherui Li}
\orcid{0009-0009-2394-4155}
\affiliation{%
  \institution{Nanyang Technological University}
  \department{College of Computing and Data Science}
  \city{Singapore}
  \country{Singapore}
}
\email{zhrli324@gmail.com}

\author{Ge Li}
\orcid{0000-0002-5828-0186}
\affiliation{%
  \institution{Peking University}
  \department{School of Computer Science}
  \city{Beijing}
  \country{China}
}
\email{lige@pku.edu.cn}

\renewcommand{\shortauthors}{Yitong Zhang, Yongmin Li, Yuetong Liu, Jia Li, Xiaoran Jia, Zherui Li, and Ge Li}

\begin{abstract}
Diffusion Large Language Models (dLLMs) have demonstrated promising capabilities and are increasingly used to produce formal languages defined by context-free grammars, such as source code and chemical expressions. However, as probabilistic models, they still struggle to generate syntactically valid outputs reliably.
A natural and promising direction to address this issue is to adapt constrained decoding techniques to enforce grammatical correctness during generation.
However, applying these techniques faces two primary obstacles. On the one hand, the non-autoregressive nature of dLLMs renders most existing constrained decoding approaches inapplicable. 
On the other hand, current approaches specifically designed for dLLMs may allow intermediate outputs that are impossible to complete into valid sentences, which significantly limits their reliability in practice. 

To address these challenges, we present \name, a constrained decoding approach specifically designed for dLLMs.
Our approach leverages a key property of dLLMs, namely their ability to predict token distributions for all positions in parallel during each forward pass.
Whenever a new token is proposed by the model, \name performs lookahead using these distributions to efficiently and reliably verify the validity of the proposed token.
This design enforces reliable constraints by preserving the potential for intermediate outputs to be extended into valid sentences.
Extensive experiments across four widely used dLLMs and \cl{five} representative benchmarks demonstrate that \name consistently outperforms existing baselines and achieves improvements in syntactic correctness, while incurring negligible runtime overhead.
\end{abstract}



\maketitle

\section{Introduction}
\label{sec:introduction}

\begin{figure}[t]
\centering
\vspace{2pt}
\begin{subfigure}[t]{0.47\textwidth}
    \centering
    \includegraphics[width=\linewidth]{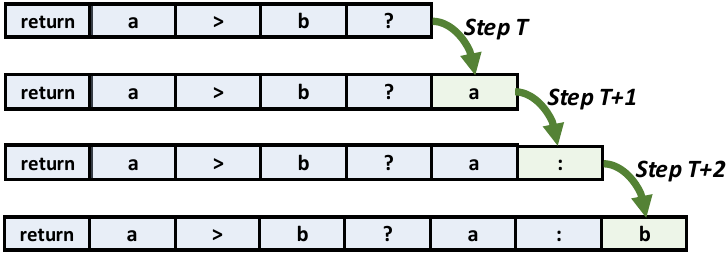}%
    \caption{Autoregressive LLM}
    \label{fig:intro-ar}
\end{subfigure}
\hfill
\begin{subfigure}[t]{0.47\textwidth}
    \centering
    \includegraphics[width=\linewidth]{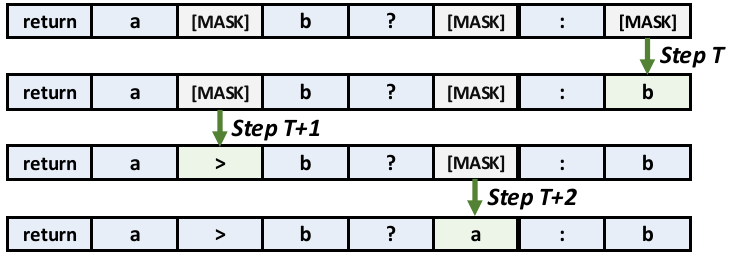}%
    \caption{Diffusion LLM}
    \label{fig:intro-dllm}
\end{subfigure}

\vspace{-0.1in}
\caption{
AR LLMs generate tokens autoregressively, so every intermediate output forms a \textit{complete prefix}, whereas dLLMs generate tokens non-autoregressively, yielding intermediate outputs that are often \textit{incomplete prefixes} due to remaining \texttt{[MASK]}.
}
\label{fig:intro}
\vspace{-0.15in}
\end{figure}

\cl{Recently, diffusion Large Language Models (dLLMs)~\cite{li2025diffuguard, liu2025wedlm, song2025seed} have attracted significant attention in both academia and industry due to their theoretical potential for high-efficiency generation~\cite{zhu2026dllm, cheng2025deer, zhang2025beyond}.}
Unlike AutoRegressive LLMs (AR LLMs), dLLMs generate tokens in a non-autoregressive manner, which allows tokens on the right to be generated before those on the left, as shown in Figure~\ref{fig:intro}. Due to their promising capabilities, they are increasingly applied to tasks requiring the generation of formal languages, which are typically defined by a Context-Free Grammar (CFG)~\cite{sun2025earley, ugare2024syncode}. For instance, a growing body of research has explored the potential of dLLMs in generating source code~\cite{diffucoder, dreamcoder} and chemical expressions~\cite{igcd}. However, as probabilistic models, \textit{\textbf{dLLMs still struggle to generate syntactically valid outputs reliably in formal languages}}~\cite{yang2025difftester, suresh2025dingo}, which limits their broader applicability in real-world scenarios. 
To assess the severity of this issue, we conducted a preliminary experiment on \dataname{HumanEval-CPP}~\cite{humaneval-x}. 
Our results show that even leading dLLMs suffer from high syntax error rates. For example, \llmname{Dream-v0-Instruct-7B}~\cite{dream} exhibits a syntax error rate of up to 23.8\%.

To address this reliability issue, a promising direction is to adapt constrained decoding techniques, which have been widely used to ensure that AR LLMs produce syntactically valid outputs~\cite{willard2023efficient, ugare2024syncode, wang2024fantastic}. These approaches typically rely on grammar parsers~\cite{aho1974lr, deremer1971simple, earley} to guarantee the validity of each newly generated token~\cite{wagner1998efficient, llguidance, willard2023efficient}. Existing parsers are generally restricted to processing \textit{complete prefixes} (\ie prefixes containing no ungenerated token), which naturally aligns with the autoregressive generation process of AR LLMs~\cite{sun2025earley, deremer1971simple, geng2023grammar}. However, dLLMs generate tokens in a non-autoregressive manner. As shown in Figure~\ref{fig:intro-dllm}, dLLMs often produce \textit{incomplete prefixes} (\ie prefixes containing some ungenerated tokens -- \texttt{[MASK]}), which renders existing constrained decoding techniques inapplicable. Consequently, a key challenge lies in \textit{\textbf{how to enable constrained decoding to function effectively on such \textit{incomplete prefixes}}}. Although recent work~\cite{igcd} has attempted to design CFG-based constrained decoding for dLLMs, it fails to ensure that the constraints are reliable. More concretely, it often leads to intermediate outputs that are impossible to complete into a valid sentence in the target language, which significantly undermines its reliability. Therefore, developing constrained decoding approaches tailored to dLLMs still remains an open problem.

To address this challenge, we propose \name, a constrained decoding approach specifically designed for dLLMs. At the core of our approach is a simple but powerful idea: \textit{\textbf{lookahead-then-verify}}.
Unlike AR LLMs, which predict a single probability distribution for the next token, dLLMs predict distributions for all positions in parallel during every forward pass~\cite{gong2024scaling, gao2025self, yang2025difftester}.
Leveraging this capability, when the model proposes a token for a masked position, \name can naturally perform lookahead by sampling tokens for the remaining masked positions within the current \textit{incomplete prefix}. This process effectively transforms the \textit{incomplete prefix} into a set of \textit{complete prefixes} that reflect the likely continuations of the model and can be directly validated by most parsers~\cite{earley, aho1974lr, deremer1971simple}. Subsequently, \name employs a grammar parser to verify whether any of these lookahead \textit{complete prefixes} can be further extended into a valid sentence in the target language. The existence of such a prefix serves as a reliable guarantee that the proposed token can preserve the potential for the output to be completed into a valid sentence. If this condition is satisfied, the proposed token is accepted; otherwise, it is rejected and the model is required to propose another token.

However, implementing this idea presents two potential hurdles.
\ding{182} First, an \textit{incomplete prefix} may contain multiple masked positions, and the vocabulary size is typically very large. As a result, determining whether a proposed token should be accepted may in principle require exploring an exponential number of lookahead \textit{complete prefixes}, which is computationally infeasible. Fortunately, our preliminary experiments in Section~\ref{sec:pre} show that sampling only a small number of lookahead prefixes is usually sufficient to decide whether a proposed token is acceptable, which reduces the runtime overhead of verification to a negligible level.
\ding{183} Second, even with efficient verification, the iterative proposal and verification process may occasionally become trapped in difficult contexts, where the model repeatedly proposes tokens that fail verification. To address this issue, \name incorporates a lightweight recovery mechanism in Section~\ref{sec:recovery} that actively adjusts the current context, allowing the model to escape such stalled states while preserving constraint reliability.

Our experimental results demonstrate that \name consistently outperforms all baselines across four widely used dLLMs and \cl{five} representative benchmarks covering diverse formal languages.
Across nearly all experimental settings, \name improves the syntactic correctness rate to almost 100\%, enhancing the reliability of the model.
Furthermore, in many scenarios, it yields improvements in functional correctness that are multiples of the gains achieved by other approaches.
In terms of cost, we find that the runtime overhead introduced by \name is nearly negligible. 
We further conduct ablation studies, which show that each component contributes positively to the overall performance of \name. 

In summary, our contributions are threefold:
\begin{itemize}[leftmargin=15pt]
\item We identify an unreliability issue in existing approaches for constraining dLLMs under CFGs, where accepted tokens may render the output impossible to complete into a valid sentence.
\item We propose \name, a constrained decoding approach that reliably enforces CFG constraints for dLLMs, ensuring that intermediate outputs remain syntactically extendable.
\item We conduct extensive experiments using four widely used dLLMs. 
The results demonstrate that \name substantially improves the reliability of dLLMs in generating CFG-compliant outputs.
\end{itemize}

\section{Background and Related Work}
\label{sec:background}

\subsection{Diffusion Large Language Models}
\label{sec:bg-1}

\cl{Diffusion LLMs have recently emerged as a promising alternative to traditional autoregressive LLMs, offering the potential for more efficient token prediction~\cite{zhang2025beyond, sahoo2024simple, li2025survey}.
Recent advances have significantly narrowed the performance gap between dLLMs and AR LLMs, paving the way for the practical deployment of dLLMs in real-world scenarios.}
Early open-source dLLMs, such as LLaDA~\cite{llada} and Dream~\cite{dream}, have demonstrated competitive performance comparable to LLaMA3-8B~\cite{touvron2023llama}. Furthermore, specialized models like Dream-Coder~\cite{dreamcoder} and DiffuCoder~\cite{diffucoder} have been designed to excel in formal languages such as source code, achieving results on par with Qwen2.5-Coder-7B~\cite{hui2024qwen2} across various code generation benchmarks. Mercury Coder~\cite{mercury} and Gemini Diffusion~\cite{gemini_diffusion} have attained capabilities rivaling those of proprietary commercial AR LLMs while offering superior inference speeds.

The inference process of mainstream dLLMs follows an iterative denoising procedure in which a fully masked sequence $\mathbf{y}^0$ is gradually transformed into the final output. Formally, a dLLM $f$ introduces a special \texttt{[MASK]} token and initializes generation with every position masked, denoted as $\mathbf{y}^0 = (\texttt{[MASK]})_{i=1}^L$, where $L$ is the maximum generation length. Given a prompt $\mathbf{p}$, the model performs up to $T$ denoising steps to produce the final sequence $\mathbf{y}^T = (y_i^T)_{i=1}^L$, which contains no remaining \texttt{[MASK]} tokens.
At each step $t \in \{1, \dots, T\}$, the model performs a single forward pass to predict the token probability distributions $P_i(\cdot \mid \mathbf{p} \oplus \mathbf{y}^{t-1})$ in parallel for all positions $i$ that are still filled with \texttt{[MASK]} tokens. A small subset of these masked positions is then decoded into non-\texttt{[MASK]} tokens according to the predicted distributions. 
Crucially, the generation order does not strictly follow a left-to-right order, which means that masks at later positions may be transformed into non-\texttt{[MASK]} tokens prior to those at earlier positions.

\subsection{Constrained Decoding}

Constrained decoding~\cite{willard2023efficient, beurer2024guiding, chen2025pre, melcer2024constrained, park2025flexible, ugare2024improving, nakshatri2025constrained, ugare2024syncode, zhang2026grammar, loula2025syntactic, geng2023grammar} is widely used to ensure that the output generated by LLMs adheres to a given grammar.
However, most existing approaches in this line of work are designed for AR LLMs and rely on their left-to-right generation order. As a result, these approaches are not directly applicable to dLLMs, whose generation paradigm is inherently non-autoregressive.
More recently, several studies have begun to explore constrained decoding tailored to dLLMs. For example, \textsc{DINGO}~\cite{suresh2025dingo} introduces a dynamic programming-based decoding procedure that enables dLLMs to generate grammar-compliant outputs. However, this approach is limited to regular grammars, which restricts its applicability in realistic scenarios that typically require more expressive context-free grammars~\cite{llguidance, earley}.

The work most closely related to ours is that of M\"undler et al.~\cite{igcd}, who propose the first constrained decoding approach that enforces CFG compliance. Their approach adopts a classical rejection sampling strategy~\cite{parys2025constrained, jiang2024rocode} to impose grammatical constraints. The key idea is to reduce the validity check for a newly proposed token to a language intersection problem. 
To illustrate, consider a model generating \textsc{C} code with the current intermediate output \colorbox{gray!10}{\textsf{for (} \texttt{[MASK] [MASK] \ldots }}. If the model proposes the token \colorbox{gray!10}{\textsf{)}} for the second masked position, their approach constructs a regular language that accepts all strings matching the pattern \colorbox{gray!10}{\textsf{for (}~$\Sigma^{\color{green}1}$~\textsf{) $\Sigma^*$}}, where $\Sigma$ denotes the vocabulary. This regular language is then intersected with the target context-free language, and the proposed token is accepted if and only if this intersection is non-empty.

\vspace{-5pt}
\begin{tcolorbox}[
    breakable,
    colback=gray!10,
    colframe=gray!50,
    title=\textbf{Definition: Reliability of Constraints}
]
\vspace{-0.08in}
In this paper, we formally define the reliability of constraints as follows~\footnotemark.
\begin{itemize}[leftmargin=5pt]
    \item \textbf{Reliable Constraint:} At every decoding step, the intermediate output remains \emph{extendable}, meaning that it can still be completed into a valid sentence in the target language.
    \item \textbf{Unreliable Constraint:} There exists at least one decoding step at which the intermediate output is \emph{not extendable}, meaning that it can no longer be completed into any valid sentence in the target language.
\end{itemize}
Since constrained decoding aims to improve the reliability of LLM-generated outputs, enforcing reliable constraints is a fundamental requirement for practical deployment.
\end{tcolorbox}
\footnotetext{Following most prior work~\cite{willard2023efficient, beurer2024guiding, ugare2024syncode} on constrained decoding for AR LLMs, we define reliability without taking the maximum generation length into account.}

To ensure computational tractability, their approach has to construct the regular language by ignoring the actual length limit of masked spans, assuming that each masked span can be filled with an arbitrary number of tokens. In the example above, the regular language will be over-approximated as \colorbox{gray!10}{\textsf{for (}~$\Sigma^{\color{red}*}$~\textsf{) $\Sigma^*$}} in practice. We argue that this over-approximation introduces constraint unreliability, as it may accept proposed tokens that render the intermediate output impossible to complete into a valid sequence.
For instance, revisiting the example above, there is only a single masked position between the parentheses, while a syntactically valid \colorbox{gray!10}{\textsf{for}} statement requires at least two semicolons, which typically correspond to two tokens. Despite this requirement, the over-approximation allows the approach to accept the proposed token \colorbox{gray!10}{\textsf{)}}, even though the resulting intermediate output can never be completed into a valid program.

\vspace{4pt}
\noindent \textbf{Summary.}
Due to the non-autoregressive nature of dLLM generation, the vast majority of existing CFG-based constrained decoding approaches are rendered inapplicable. Moreover, the current approach tailored to dLLMs fails to enforce reliable constraints during inference, which substantially undermines its reliability. This highlights the need for a reliable constrained decoding approach tailored to dLLMs.

\section{Methodology}
\label{sec:method}

In this section, we propose \name, a reliable constrained decoding approach designed for dLLMs. We first provide an overview of \name in Section~\ref{sec:overview}, followed by detailed descriptions of its key components and implementation details in Section~\ref{sec:check} and Section~\ref{sec:recovery}.
\cl{Finally, we analyze the computational complexity of \name in Section~\ref{sec:complexity}.}

\begin{cp}



\subsection{Overview of \name}
\label{sec:overview}
We constrain the decoding process of dLLMs through an iterative proposal and verification mechanism, as illustrated in Algorithm~\ref{alg}. During inference, the dLLM is allowed to propose new tokens for any masked positions according to its native decoding strategy (Line 5 in Algorithm~\ref{alg}). Each proposed token updates the current output by replacing a \texttt{[MASK]} token with a non-\texttt{[MASK]} token (Line 6 in Algorithm~\ref{alg}). The effectiveness of this decoding process critically depends on how the validity of each proposed token is assessed. 
To this end, we apply \textbf{Lookahead-Based Verification} (Section~\ref{sec:check}), the core component of \name, to reliably determine whether a proposed token should be accepted or rejected (Lines 7--13 in Algorithm~\ref{alg}). If the token is accepted, the dLLM proceeds to generate the next token; otherwise, it attempts to propose an alternative.

In certain cases, the model may repeatedly fail to produce an acceptable proposal~\cite{igcd, jiang2024rocode, ugare2024itergen, melcer2024constrained}. To mitigate this issue, we track the number of consecutive failed proposal attempts (Lines 10 and 12) and invoke \textbf{Cache-Enhanced Recovery} (Section~\ref{sec:recovery}) once this number reaches a predefined threshold (Lines 14--17 in Algorithm~\ref{alg}), enabling the model to recover from stalled inference states.

This process continues until no \texttt{[MASK]} token remains (Line 4 in Algorithm~\ref{alg}), after which \name returns the final output (Lines 18--19 in Algorithm~\ref{alg}).
\end{cp}

\begin{algorithm}[t]
\small
\setlength{\baselineskip}{0.97\baselineskip}
\caption{\name}
\label{alg}
\begin{algorithmic}[1]
\Require dLLM model $f$, Context-Free Grammar $\mathcal{G}$, prompt $\mathbf{p}$, lookahead size $N$, attempt budget $\tau$
\State Initialize $\mathbf{y}$ with \texttt{[MASK]} tokens
\State Initialize counter $c_{\text{fail}} \leftarrow 0$
\State Initialize $\mathbf{y}_{\text{cache}} \leftarrow \epsilon$

\While{there are remaining \texttt{[MASK]} tokens}
    \State Propose a token $t^*$ for a masked position $i^*$ using $f$\;
    \State Create a temporary updated output $\mathbf{y}' \leftarrow \mathbf{y}$ with $\mathbf{y}'_{i^*} \leftarrow t^*$

    \If{\textbf{Lookahead-Based Verification}($\mathbf{y}'$) == \texttt{true}}
    \Comment{\textit{\textbf{Section~\ref{sec:check}}}}
        \State $\mathbf{y} \leftarrow \mathbf{y}'$
        \Comment{\textit{Accept the proposed token}}
        \State Update $\mathbf{y}_{\text{cache}}$ with a verified \textit{complete prefix}
        \State $c_{\text{fail}} \leftarrow 0$
    \Else
        \State $c_{\text{fail}} \leftarrow c_{\text{fail}} + 1$
        \Comment{\textit{Reject the proposed token}}
    \EndIf

    \If{$c_{\text{fail}} \ge \tau$} 
    \Comment{\textit{\textbf{Section~\ref{sec:recovery}}}}
        \State Apply \textbf{Cache-Enhanced Recovery} to $\mathbf{y}$ with $\mathbf{y}_{\text{cache}}$\;
        \State $c_{\text{fail}} \leftarrow 0$
    \EndIf
\EndWhile
\State \textbf{Output:} final output $\mathbf{y}$ without \texttt{[MASK]} tokens
\end{algorithmic}
\end{algorithm}

\subsection{Lookahead-Based Verification}
\label{sec:check}

\begin{figure*}[t]
    \centering
    \includegraphics[width=0.98\textwidth]{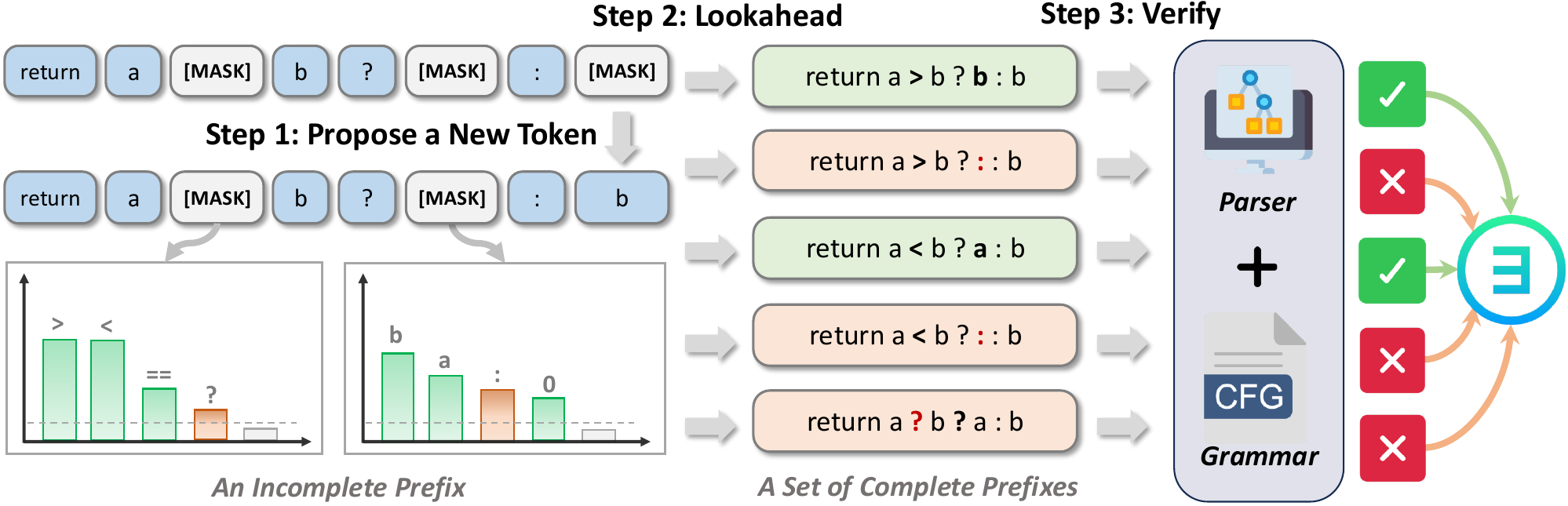} 
    \caption{The dLLM proposes a token for one masked position, which is subsequently verified through Lookahead-Based Verification. A proposed token is accepted if at least one lookahead \textit{complete prefix} is extendable under the target grammar; otherwise, it is rejected.}
    \label{fig:method}
    \vspace{-0.15in}
\end{figure*}

In this section, we introduce Lookahead-Based Verification, which is the core of \name. Figure~\ref{fig:method} illustrates how verification is performed.

When the model proposes a new token, it updates the current output by replacing a specific \texttt{[MASK]} token with a non-\texttt{[MASK]} token.
The goal of Lookahead-Based Verification is to reliably verify whether this update preserves the potential for the output to be completed into a valid sentence.

\subsubsection{Problem Formulation.}
Formally, let $\mathbf{y} \in (\Sigma \cup \{\texttt{[MASK]}\})^{L}$ denote the updated output that includes the newly proposed token.
We define the index of the rightmost non-\texttt{[MASK]} token as
\begin{equation}
r(\mathbf{y}) = \max \{\, i \mid y_i \ne \texttt{[MASK]} \,\},
\end{equation}
where $r(\mathbf{y}) = 0$ if all tokens are masked.
This index determines a prefix $\mathbf{y}^p = \mathbf{y}_{1:r(\mathbf{y})}$, which is typically an \textit{incomplete prefix} containing several \texttt{[MASK]} tokens.
We denote the set of masked positions inside this prefix as
\begin{equation}
\mathcal{M}(\mathbf{y}^p) = \{\, i \mid 1 \le i < r(\mathbf{y}),\ y^p_i = \texttt{[MASK]} \,\}.
\end{equation}

A proposed token can be reliably accepted as long as there exists at least one assignment of tokens to these remaining masked positions such that the resulting \textit{complete prefix} is a valid prefix:
\begin{equation}
\exists \mathbf{z} \in \Sigma^{|\mathcal{M}(\mathbf{y}^p)|}
\quad \text{s.t.} \quad
\mathsf{IsExtendable}\!\left(\Phi(\mathbf{y}^p, \mathbf{z})\right) = \texttt{true},
\label{eq:condition}
\end{equation}
where $\Phi(\cdot, \cdot)$ is a function that fills the masked positions in $\mathbf{y}^p$ using the assignment $\mathbf{z}$. The function $\mathsf{IsExtendable}(\cdot)$ is a standard capability provided by most grammar parsers (\eg Earley~\cite{earley} and LR~\cite{aho1974lr} parsers) and can efficiently check whether a \textit{complete prefix} can be extended into a valid sentence.

\subsubsection{Motivation and Feasibility.}
\label{sec:pre}
dLLMs offer a distinct advantage over AR LLMs: they predict token distributions for all positions simultaneously.
This property enables a natural \textit{lookahead-then-verify} strategy.
Given an \textit{incomplete prefix} $\mathbf{y}^p$, we can perform lookahead by filling the masked positions in $\mathcal{M}(\mathbf{y}^p)$ according to the model's predicted distributions, thereby transforming $\mathbf{y}^p$ into a set of \textit{complete prefixes} that reflect the model's likely continuations.
These complete prefixes can then be checked using a grammar parser via $\mathsf{IsExtendable}(\cdot)$, allowing us to verify whether the current output reliably remains extendable.

The effectiveness and efficiency of this strategy depend on how many \textit{complete prefixes} need to be sampled.
While sampling a larger number of \textit{complete prefixes} increases the likelihood of acceptance, it may also introduce non-negligible overhead.
To assess this trade-off and validate the feasibility of our approach, we conducted an empirical study.

\begin{figure}[t]
\centering
\vspace{2pt}
\begin{subfigure}[t]{0.47\textwidth}
    \centering
    \includegraphics[width=\linewidth]{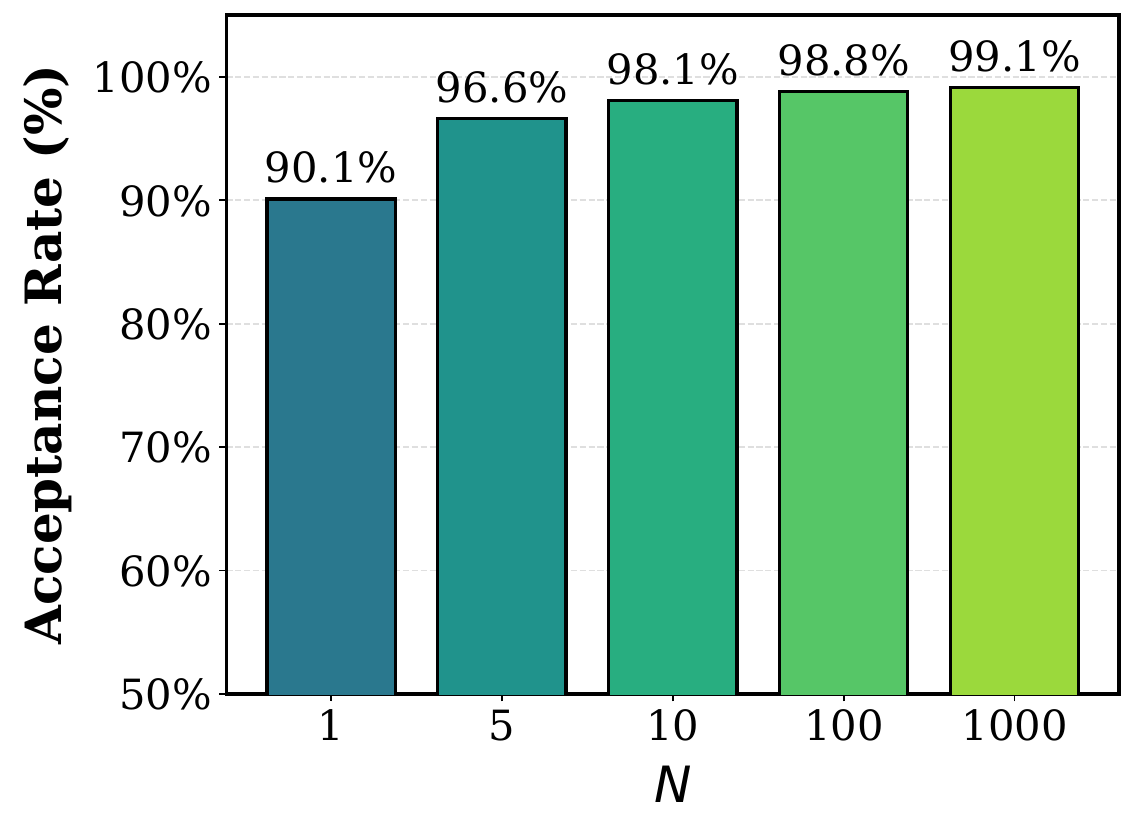}%
    \caption{Dream-v0-Instruct-7B}
    \label{fig:pre1}
\end{subfigure}
\hfill
\begin{subfigure}[t]{0.47\textwidth}
    \centering
    \includegraphics[width=\linewidth]{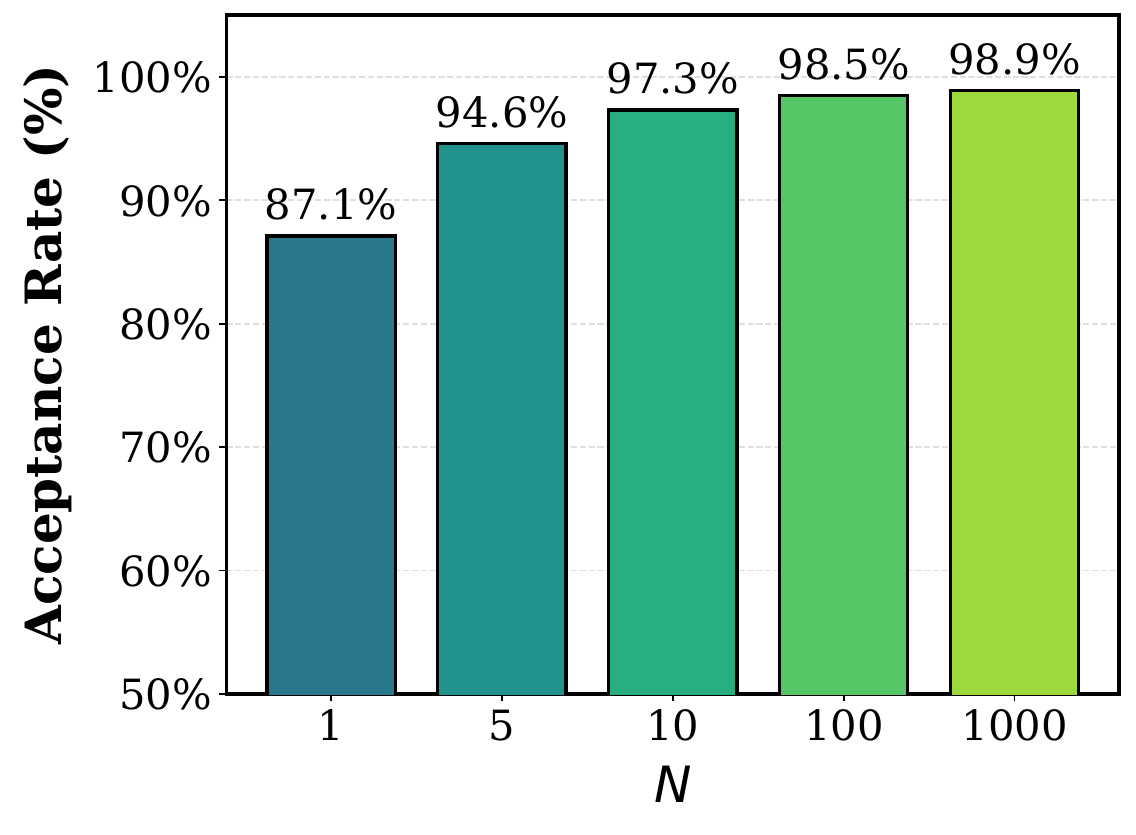}%
    \caption{LLaDA-8B-Instruct}
    \label{fig:pre2}
\end{subfigure}

\caption{
Experimental results from our preliminary empirical study. 
}
\label{fig:preliminary-results}
\vspace{-0.15in}
\end{figure}

We evaluated two widely used models, Dream-v0-Instruct-7B~\cite{dream} and LLaDA-8B-Instruct~\cite{llada}, on the \dataname{HumanEval-CPP} benchmark~\cite{humaneval-x}.
For each test instance, we randomly selected a prefix from the reference solution and used it to prefill the output, thereby forming a \textit{complete prefix}.
To simulate an intermediate inference state, we then randomly masked tokens within this \textit{complete prefix} to obtain an \textit{incomplete prefix}, which was guaranteed to be extendable. Specifically, each token was retained with a probability of 80\% and replaced with \texttt{[MASK]} otherwise.
We then performed a single forward pass to obtain the predicted probability distributions at each masked position. To test the feasibility of our approach, we sampled $N$ distinct \textit{complete prefixes} from these distributions and checked whether any of them could be extended into a valid sentence under the target CFG.
A higher acceptance rate indicates that our motivation is practically feasible in reliably determining the extendability of \textit{incomplete prefixes}.
As illustrated in Figure~\ref{fig:preliminary-results}, we observe that sampling only a small number of \textit{complete prefixes} is typically sufficient to uncover a valid \textit{complete prefix} whenever the underlying \textit{incomplete prefix} has the potential to be extended into one valid sentence. In particular, with a lookahead size of $N=10$, the acceptance rate reaches 98.1\% for \llmname{Dream-v0-Instruct-7B} and 97.3\% for \llmname{LLaDA-8B-Instruct}. 
These results indicate that the proposed \textit{lookahead-then-verify} strategy is feasible in practice.
 
\subsubsection{Verification Procedure.}
Based on the feasibility of the proposed \textit{lookahead-then-verify} strategy, we now describe how it is implemented in practice.

Let $P_j(\cdot)$ denote the predicted token distribution at each masked position $j \in \mathcal{M}(\mathbf{y}^p)$.
To perform \textbf{lookahead}, we sample $N$ distinct assignments $\{\mathbf{z}^{(1)}, \ldots, \mathbf{z}^{(N)}\}$ according to these marginal distributions $P_j(\cdot)$. Formally, each assignment $\mathbf{z}$ is sampled from the joint distribution:
\begin{equation}
P(\mathbf{z}) = \prod_{j \in \mathcal{M}(\mathbf{y}^p)} P_j(z_j).
\end{equation}
Each sampled assignment $\mathbf{z}^{(n)}$ is then used to fill all masked positions in $\mathbf{y}^{p}$, resulting in a \textit{complete prefix}:
\begin{equation}
\mathbf{s}^{(n)} = \Phi(\mathbf{y}^{p}, \mathbf{z}^{(n)}), n \in \{1, \cdots, N\},
\label{eq:candidate}
\end{equation}
which no longer contains any \texttt{[MASK]} tokens.

In the \textbf{verification} stage, we apply grammar parsers to each \textit{complete prefix} to determine whether it is extendable in parallel.
The proposed token is accepted if and only if at least one \textit{complete prefix} satisfies:
\begin{equation}
\exists n \in \{1,\ldots,N\}
\quad \text{s.t.} \quad
\mathsf{IsExtendable}(\mathbf{s}^{(n)}) = \texttt{true}.
\label{eq:validation}
\end{equation}
Such a \textit{complete prefix} serves as a constructive witness that the updated output remains grammatically extendable, thereby naturally ensuring the reliability of the constraint. Conversely, if no valid \textit{complete prefix} is found after $N$ lookahead attempts, the proposed token is rejected, and another proposed token is sampled from the model's original predicted distribution.

\subsection{Cache-Enhanced Recovery}
\label{sec:recovery}

In practice, the iterative proposal and verification process may occasionally enter a challenging state in which the current context makes it difficult for the model to produce an acceptable token. Concretely, the model may repeatedly propose tokens that fail the Lookahead-Based Verification, causing the decoding process to nearly stall. To address this issue, we introduce a recovery strategy that \textbf{actively modifies the current context}.

Our recovery strategy leverages a byproduct of the Lookahead-Based Verification process described in Section~\ref{sec:check}. Specifically, whenever a \textit{complete prefix} $\mathbf{s}^{(n)}$ passes the verification, it constitutes a valid prefix that is both syntactically extendable and favored by the model. We record such prefixes by maintaining a global variable $\mathbf{y}_{\text{cache}}$, which is updated whenever a verification succeeds:
\begin{equation}
\mathbf{y}_{\text{cache}} \leftarrow \mathbf{s}^{(n)}, \quad \text{if } \mathsf{IsExtendable}(\mathbf{s}^{(n)}) = \texttt{true}.
\end{equation}

During inference, we monitor the number of consecutive failed proposal attempts. When this count reaches a predefined proposal budget $\tau$, we infer that the model is trapped in a difficult context. To break this impasse, we replace the prefix of the current output with $\mathbf{y}_{\text{cache}}$:
\begin{equation}
\mathbf{y}_{1:r(\mathbf{y})} \leftarrow \mathbf{y}_{\text{cache}}.
\end{equation}

However, when $\mathbf{y}_{1:r(\mathbf{y})}$ contains no remaining \texttt{[MASK]} tokens (\ie $\mathbf{y}_{1:r(\mathbf{y})}$ is identical to $\mathbf{y}_{\text{cache}}$), this replacement alone does not change the current context. To ensure that the context can be effectively modified, we generate one additional token following the current prefix.
Many parsers used in existing constrained decoding approaches~\cite{llguidance, chen2025pre, wagner1998efficient} can efficiently compute the set of valid successor tokens given a \textit{complete prefix}. All tokens in this set are guaranteed to preserve the syntactic validity of the prefix. Let $\mathcal{V} = \mathsf{NextTokens}(\mathbf{y}_{\text{cache}})$ denote the set of such admissible tokens. We define an adjusted probability distribution for the next position as follows:
\begin{equation}
\hat{P}_{r(\mathbf{y}) + 1}(t) \propto P_{r(\mathbf{y}) + 1}(t)\,\mathbb{I}[t \in \mathcal{V}].
\end{equation}
We then sample the next token from $\hat{P}_{r(\mathbf{y}) + 1}$ and continue inference with the standard proposal and verification process, thereby preserving both the non-autoregressive generation behavior and the constraint reliability of subsequent generation.

\begin{cp}
\subsection{Complexity Analysis}
\label{sec:complexity}
According to Algorithm~\ref{alg}, each decoding step allows at most $\tau$ failed proposals, and each proposal requires at most $N$ lookahead verifications. With an Earley parser~\cite{earley}, verifying a prefix of length $\ell$ takes $O(\ell^3)$ time for a general CFG. Therefore, the additional cost of generating one accepted token is bounded by $O(\tau N \ell^3)$. For practical deterministic CFGs, this cost reduces to $O(\tau N \ell)$.
\end{cp}

\section{Experimental Setup}
\label{sec:setup}
In this section, we introduce the experimental setup, including the benchmarks and their corresponding context-free grammars, the compared baselines, the evaluation metrics, the models, and other implementation details. Based on this setup, we conduct a series of experiments designed to address the following Research Questions (RQs).

\begin{itemize}[leftmargin=15pt]
    \item \textbf{RQ1}: How well does \name ensure the syntactic correctness of dLLM outputs compared to baseline approaches?
    \item \textbf{RQ2}: How well does \name improve the functional correctness of dLLM outputs?
    \item \cl{\textbf{RQ3}: What runtime overhead does \name introduce?}
    \item \textbf{RQ4}: How does each component of \name contribute to the overall performance?
    \item \textbf{RQ5}: How do the hyperparameters affect the performance of \name?
\end{itemize}

\subsection{Benchmark}
\label{sec:benchmark}
To ensure a fair comparison, we follow existing studies~\cite{igcd} in selecting benchmarks for evaluation. Specifically, we adopt \cl{five} benchmarks that require the model to generate representative formal languages defined by context-free grammars. 

\begin{itemize}[leftmargin=12pt]
    \begin{cp}
    \item \textbf{Code Generation:} To evaluate the generalizability of our approach across different programming languages, we consider three code generation benchmarks derived from HumanEval~\cite{humaneval}, namely \textbf{HumanEval-CPP}, \textbf{HumanEval-Java}, and \textbf{HumanEval-Go}~\cite{humaneval-x}, which are designed to evaluate the code generation ability of LLMs in C++, Java, and Go, respectively. Each benchmark contains 164 problems. For each problem, the dataset provides a natural language problem description and a set of test cases to verify the correctness of the generated output. We include these benchmarks to demonstrate the performance of \name in code generation scenarios across multiple programming languages. In the following sections, we refer to these benchmarks as \dataname{CPP-Bench}, \dataname{JAVA-Bench}, and \dataname{GO-Bench}, respectively.  
    \end{cp}

    \item \textbf{JSON Generation:} \textbf{JSON-Mode-Eval-Extended~\cite{igcd}} is constructed based on the JSON-Mode-Eval benchmark~\cite{nous_json_mode_eval} to evaluate the information extraction capabilities of LLMs using the JSON format. The benchmark consists of 272 problems. Each problem comprises a natural language snippet, a JSON Schema (corresponding to a specific CFG), and a ground truth JSON response. We include this benchmark because it can effectively reflect the performance of \name in generating JSON, a formal language that is widely employed in information extraction and agent function calling tasks. In the following sections, we refer to this benchmark as \dataname{JSON-Bench}.
    
    \item \textbf{Chemical Expression Generation:} \textbf{SMILES-Eval~\cite{igcd}} is constructed to assess LLM capabilities in generating SMILES chemical molecule representations from natural language descriptions. The benchmark consists of 167 problems. Each problem includes a natural language molecule description and a corresponding SMILES string as the ground truth. We include this benchmark to highlight the applicability of \name in scientific domains. In the following sections, we refer to this benchmark as \dataname{SMILES-Bench}.
\end{itemize}

\begin{cp}
\subsection{Grammar}
\label{sec:grammar}
For CFG-constrained decoding, a concrete grammar specification is required. For \dataname{CPP-Bench}, \dataname{JAVA-Bench}, \dataname{GO-Bench}, and \dataname{SMILES-Bench}, we construct grammars based on the standard syntax of C++, Java, Go, and SMILES, respectively. For \dataname{JSON-Bench}, where each problem is associated with a distinct JSON Schema, we follow prior work~\cite{igcd} and automatically construct a corresponding grammar from the provided JSON Schema for each problem.
\end{cp}

\subsection{Baselines}
\label{sec:baseline}
We compare \name against three baseline approaches. \ding{182} First, we include the constrained decoding approach proposed by M\"undler et al.~\cite{igcd}, which we refer to as \baselinename{IG-CD} (Intersection-Guided Constrained Decoding). \ding{183} Second, we include the unconstrained generation setting, denoted as \baselinename{NO-CD}. \ding{184} In addition, we design a straightforward constrained decoding baseline that enforces a strictly left-to-right generation order for dLLMs, thereby allowing the direct application of constrained decoding techniques~\cite{llguidance} originally developed for AR LLMs. We refer to this baseline as \baselinename{FS-CD} (Forced-Sequential Constrained Decoding).

\subsection{Metrics}
\label{sec:metric}

We evaluate all approaches using three metrics: \textit{syntactic@k}, \textit{functional@k}, and average inference time.  

\ding{182} \textit{\textbf{syntactic@k}} measures the proportion of problems for which \textbf{at least one} of the $k$ generated outputs satisfies the syntactic constraints specified by the given context-free grammar.

\ding{183} \textit{\textbf{functional@k}} measures the proportion of problems for which \textbf{at least one} of the $k$ generated outputs exhibits correct functional behavior. \cl{For \dataname{CPP-Bench}, \dataname{GO-Bench}, and \dataname{JAVA-Bench}}, functional correctness is determined by executing the provided test cases. For the other benchmarks, it is assessed by exact matching against the ground-truth outputs. In general, syntactic correctness is a prerequisite for functional correctness.

\ding{184} \textbf{Average Inference Time} records the average decoding time per generation, providing a measure of the overhead introduced by different constrained decoding approaches.

\subsection{Models}
\label{sec:models}
We evaluate our approach using four widely used dLLMs:

\begin{itemize}[leftmargin=15pt]
    \item \llmname{LLaDA-8B-Instruct}~\cite{llada} is released by researchers from Renmin University of China. It is an early open-source diffusion large language model with 8 billion parameters and is trained entirely from scratch. In the following, we refer to this model as \llmname{LLaDA-8B}.
    \item \llmname{LLaDA-1.5}~\cite{llada15} is also released by researchers from Renmin University of China. It extends \llmname{LLaDA-8B-Instruct} by incorporating Variance Reduced Preference Optimization (VRPO) during training. Compared with \llmname{LLaDA-8B-Instruct}, it achieves improved performance across a wide range of tasks, including code-related tasks. 
    \item \llmname{Dream-v0-Instruct-7B}~\cite{dream} is released by researchers from The University of Hong Kong. It is initialized from an autoregressive language model and demonstrates strong overall capabilities. In the following, we refer to this model as \llmname{Dream-7B}.
    \item \llmname{DiffuCoder-7B-cpGRPO}~\cite{diffucoder} is released by researchers from Apple and The University of Hong Kong. It is post-trained with coupled-GRPO on 21K code examples, leading to substantial improvements in code generation performance. In the following, we refer to this model as \llmname{DiffuC-7B}.
\end{itemize}

\subsection{Implementation Details}
\label{sec:implementation}
Following prior studies~\cite{yang2025difftester, igcd}, we apply a simple and commonly used optimization for all approaches: once a dLLM generates an \texttt{[EOS]} token, all subsequent token positions are set to \texttt{[EOS]} to accelerate inference.
For all experiments, we adopt the widely used semi-autoregressive strategy~\cite{llada, arriola2025block, yang2025mmada, d2f, tracerl} and set the block size to 32, as is common in prior studies~\cite{llada15, nguyen2025attention, hu2025accelerating}.
Additionally, for each experiment, we conduct at least four independent runs and report the average values of the evaluation metrics.
For both \baselinename{FS-CD} and \name, we employ the Earley algorithm~\cite{earley} to implement the required grammar parser, which allows us to handle arbitrary context-free grammars.
All experiments are conducted on a server equipped with eight NVIDIA A100 GPUs, each with 40 GB of memory, and two Intel Xeon Gold 6348 CPUs, with 28 cores per socket and 56 physical cores in total.

Regarding the hyperparameters of the dLLMs, we set the maximum generation length $L$ to 256. The number of denoising steps $T$ is set to 128, which provides a balanced trade-off between model effectiveness and efficiency~\cite{zhang2025beyond, llada}. The temperature is fixed to 0.2 for all experiments.
Regarding the hyperparameters specific to \name, we set the lookahead size $N$ defined in Section~\ref{sec:check} to 10 and the attempt budget $\tau$ defined in Section~\ref{sec:recovery} to 5.

\section{Experimental Results}
\label{sec:result}

\subsection{RQ1: How Well Does \name Ensure the Syntactic Correctness of dLLM Outputs?}
\label{sec:rq1}

\textbf{Motivation.}
The primary objective of \name is to ensure that the outputs generated by dLLMs reliably conform to the syntax defined by a CFG. 
To assess this capability, we compare the syntactic correctness of the outputs generated by our approach with that of the baseline outputs in this RQ.

\begin{table}[t]
\cc
\centering
\small
\vspace{2pt}
\renewcommand{\arraystretch}{0.8}
\setlength{\tabcolsep}{2.8pt}
\caption{\cc \textit{syntactic@k} (\%) on \dataname{JSON-Bench}, \dataname{CPP-Bench}, \dataname{JAVA-Bench}, \dataname{GO-Bench}, and \dataname{SMILES-Bench}, where $k\in\{1,5,10\}$. 
The \textbf{best result} for each model, benchmark, and $k$ is highlighted in bold.}
\label{tab:rq1}
\vspace{-0.07in}
\scalebox{0.97}{
\begin{tabular}{c ccc ccc ccc ccc ccc}
\toprule
\multirow{2}{*}{\textbf{Method}} &
\multicolumn{3}{c}{\textbf{\dataname{JSON-Bench}}} &
\multicolumn{3}{c}{\textbf{\dataname{CPP-Bench}}} &
\multicolumn{3}{c}{\textbf{\dataname{JAVA-Bench}}} &
\multicolumn{3}{c}{\textbf{\dataname{GO-Bench}}} &
\multicolumn{3}{c}{\textbf{\dataname{SMILES-Bench}}} \\ 
\cmidrule(lr){2-4}\cmidrule(lr){5-7}\cmidrule(lr){8-10}\cmidrule(lr){11-13}\cmidrule(lr){14-16}
& \textbf{k=1} & \textbf{k=5} & \textbf{k=10}
& \textbf{k=1} & \textbf{k=5} & \textbf{k=10}
& \textbf{k=1} & \textbf{k=5} & \textbf{k=10}
& \textbf{k=1} & \textbf{k=5} & \textbf{k=10}
& \textbf{k=1} & \textbf{k=5} & \textbf{k=10}\\
\midrule

\multicolumn{16}{c}{\textbf{\llmname{LLaDA-8B}}} \\
\arrayrulecolor{gray}\midrule\arrayrulecolor{black}
\baselinename{NO-CD} 
& 85.4  & 90.1 & 96.4 
& 74.2  & 94.1 & 95.6 

& 72.0 & {86.6} & {89.6}
& 72.0 & {86.1} & {87.5}

& 66.0  & 75.9  & 78.9  
\\
\baselinename{FS-CD} & 
{94.9} & {96.4} & 96.4 
& 66.3 & 66.7 & 66.7

& {78.7} & 79.4 & 79.4
& {81.7} & 82.2 & 82.3

& 75.7 & 75.9  & 76.1  
\\
\baselinename{IG-CD} 
& 94.8 & 95.7 & {96.7} 
& {86.1} & {96.7} & {98.1} 

& -- & -- & -- 
& -- & -- & --

& {94.3} & {98.7}  & {99.2}  
\\
\rowcolor[HTML]{F3F3F3}
\name                
& \textbf{99.5} & \textbf{99.6} & \textbf{99.6} 
& \textbf{90.9} & \textbf{99.3} & \textbf{99.4} 

& \textbf{92.1} & \textbf{94.5} & \textbf{95.1}
& \textbf{91.5} & \textbf{97.7} & \textbf{99.4}

& \textbf{98.7} & \textbf{100.0} & \textbf{100.0} 
\\
\midrule

\multicolumn{16}{c}{\textbf{\llmname{LLaDA-1.5}}} \\
\arrayrulecolor{gray}\midrule\arrayrulecolor{black}
\baselinename{NO-CD} 
& 86.0  & 87.7 & 87.9 
& 78.2  & 92.0 & 94.8 

& {78.1} & {87.2} & {90.9}
& 55.5 & 71.3 & 76.8

& 62.3  & 77.7 & 83.1 
\\
\baselinename{FS-CD} 
& {96.0} & {97.7} & {98.1} 
& 73.8 & 74.2 & 74.2 

& 78.1 & 78.1 & 78.1
& {77.4} & {82.3} & {83.5}

& 78.3 & 78.5 & 78.5 
\\
\baselinename{IG-CD} 
& 94.0  & 95.4 & 96.2 
& {89.5} & {97.3} & \textbf{98.8} 

& -- & -- & --
& -- & -- & --

& {93.7} & {97.0} & {97.6} 
\\
\rowcolor[HTML]{F3F3F3}
\name                
& \textbf{99.7}  & \textbf{99.8} & \textbf{99.8} 
& \textbf{91.3}  & \textbf{98.8} & \textbf{98.8} 

& \textbf{91.5} & \textbf{94.5} & \textbf{95.1}
& \textbf{89.6} & \textbf{94.5} & \textbf{97.0}

& \textbf{96.6}  & \textbf{100.0} & \textbf{100.0} 
\\
\midrule

\multicolumn{16}{c}{\textbf{\llmname{Dream-7B}}} \\
\arrayrulecolor{gray}\midrule\arrayrulecolor{black}
\baselinename{NO-CD} 
& 86.8 & 89.3 & 89.3 
& 76.2 & 79.2 & 81.1 

& 69.5 & 70.1 & 70.1
& 64.6 & 70.7 & 72.0

& 74.0 & 76.2  & 76.8  
\\
\baselinename{FS-CD} 
& {96.8} & {98.2} & {98.2} 
& 83.1 & 84.4 & 84.4 

& {84.8} & {85.1} & {85.1}
& {78.4} & {78.7} & {78.7}

& 93.1 & 93.8  & 93.8  
\\
\baselinename{IG-CD} 
& 94.9  & 95.5 & 95.6 
& {87.6} & {90.4} & {90.9} 

& -- & -- & -- 
& -- & -- & --

& {95.1} & {96.7}  & {97.4} 
\\
\rowcolor[HTML]{F3F3F3}
\name                
& \textbf{99.0} & \textbf{99.6} & \textbf{99.7}
& \textbf{96.0} & \textbf{99.4} & \textbf{99.4} 

& \textbf{90.9} & \textbf{93.3} & \textbf{94.5}
& \textbf{89.3} & \textbf{96.3} & \textbf{98.2}

& \textbf{99.4} & \textbf{100.0} & \textbf{100.0} 
\\
\midrule

\multicolumn{16}{c}{\textbf{\llmname{DiffuC-7B}}} \\
\arrayrulecolor{gray}\midrule\arrayrulecolor{black}
\baselinename{NO-CD} 
& 89.9 & 90.4  & 90.4 
& 79.9 & 80.5 & 81.7  

& 82.3 & 82.9 & 83.5
& 78.7 & 79.3 & 79.9

& 70.1 & 71.3 & 71.3 
\\
\baselinename{FS-CD}
& {96.9} & {98.1}  & {98.1} 
& {93.8} & 94.0 & 94.3  

& {91.5} & {91.5} & {91.5}
& {92.1} & {92.5} & {92.7}

& 96.0 & 96.3 & 96.3 
\\
\baselinename{IG-CD} 
& 92.1  & 92.7  & 92.7 
& {93.8} &{95.1} & {95.3}  

& -- & -- & -- 
& -- & -- & --

& {96.4} & {96.6} & {96.7} 
\\
\rowcolor[HTML]{F3F3F3}
\name                
& \textbf{99.6} & \textbf{100.0}  & \textbf{100.0} 
& \textbf{97.0} & \textbf{99.6} & \textbf{100.0} 

& \textbf{93.3} & \textbf{95.5} & \textbf{95.7}
& \textbf{96.5} & \textbf{99.1} & \textbf{99.4}

& \textbf{98.8} & \textbf{99.3} & \textbf{99.4} 
\\
\bottomrule
\end{tabular}
}
\vspace{-0.15in}
\end{table}

\begin{cp}
\vspace{3pt}
\noindent \textbf{Setting.}
We apply the baselines and \name to the four models described in Section~\ref{sec:models} and evaluate their performance on the five benchmarks introduced in Section~\ref{sec:benchmark}. We use \textit{syntactic@k} as the evaluation metric with $k \in \{1, 5, 10\}$, as defined in Section~\ref{sec:metric}. 
Unfortunately, we cannot evaluate \baselinename{IG-CD} on \dataname{GO-Bench} or \dataname{JAVA-Bench}, because the original paper~\cite{igcd} does not provide IG-CD-compatible grammars for Go and Java, nor does it publicly release the grammar construction rules used by IG-CD. Therefore, we cannot faithfully reproduce IG-CD on these two benchmarks.
\end{cp}

\vspace{3pt}
\noindent \textbf{Results.}
The \textit{syntactic@k} scores of different approaches are reported in Table~\ref{tab:rq1}.

\textbf{\name demonstrates superior performance in ensuring that dLLM outputs conform to CFG constraints.} Compared to the unconstrained baseline, \name substantially improves syntactic correctness, reaching close to 100\% in most settings. For example, on the syntactically simpler \dataname{SMILES-Bench} with \llmname{LLaDA-8B}, \name raises \textit{syntactic@1} from 66.0\% to 98.7\%, \textit{syntactic@5} from 75.9\% to 100.0\%, and \textit{syntactic@10} from 78.9\% to 100.0\%. On the more syntactically complex \dataname{CPP-Bench}, our approach also achieves strong performance. With \llmname{DiffuC-7B}, \name increases \textit{syntactic@1} from 79.9\% to 97.0\%, \textit{syntactic@5} from 80.5\% to 99.6\%, and \textit{syntactic@10} from 81.7\% to 100.0\%.
\cl{We further observe similarly strong improvements on the relatively lower-resource \dataname{GO-Bench}. For example, \name raises \textit{syntactic@10} from 87.5\% to 99.4\% on \llmname{LLaDA-8B}, and from 72.0\% to 98.2\% on \llmname{Dream-7B}. The only exception is \dataname{JAVA-Bench}. Although \name still yields substantial improvements on this benchmark, its syntactic correctness remains harder to push to nearly 100\%. By examining representative cases, we find that this limitation mainly arises because Java programs are typically more verbose, making the max generation length of 256 occasionally insufficient, rather than reflecting a weakness of our approach itself.}
These results demonstrate that \name greatly enhances the reliability of dLLM outputs when generating formal languages such as C++, \cl{Java, Go, JSON,} and SMILES.

\textbf{\name outperforms existing constrained decoding approaches across most evaluated benchmarks and models.} For example, on \dataname{JSON-Bench} with \llmname{LLaDA-8B}, \name improves \textit{syntactic@1} over the best baseline \baselinename{FS-CD} by 4.6\%. On \dataname{CPP-Bench} with \llmname{Dream-7B}, \name outperforms the best baseline \baselinename{IG-CD} by 8.4\% in terms of \textit{syntactic@1}. In addition, we observe that \name benefits more from increasing $k$ than FS-CD. For instance, on \dataname{CPP-Bench} with \llmname{LLaDA-1.5}, as $k$ increases from 1 to 10, \name improves \textit{syntactic@k} from 91.3\% to 98.8\%, whereas FS-CD shows only a marginal increase from 73.8\% to 74.2\%.
\cl{A similar trend is even more evident on \dataname{JAVA-Bench} with \llmname{DiffuC-7B}: as $k$ increases from 1 to 10, \baselinename{FS-CD} shows no improvement at all, whereas \name improves monotonically throughout.}
We attribute this behavior to the fact that our approach preserves the non-autoregressive generation mechanism of dLLMs, which provides greater flexibility and diversity during inference.

\begin{figure*}[!t]
    \vspace{2pt}
    \centering
    \includegraphics[width=0.9\textwidth]{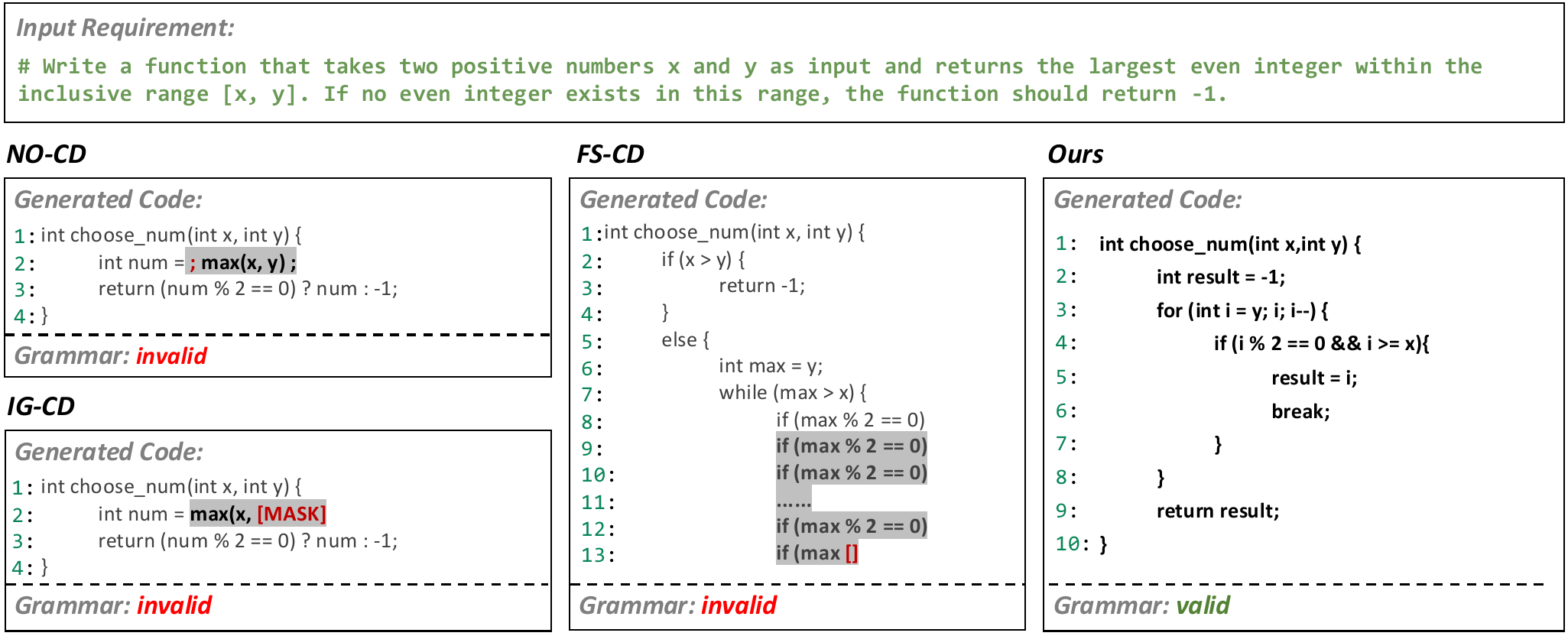} 
    \vspace{-5pt}
    \caption{
    Outputs generated by \baselinename{NO-CD}, \baselinename{FS-CD}, \baselinename{IG-CD}, and \name on \llmname{Dream-7B} for a problem from \dataname{CPP-Bench}.}
    \label{fig:rq1}
    \vspace{-0.15in}
\end{figure*}

We further conduct a case study to analyze why \name outperforms existing approaches. Figure~\ref{fig:rq1} presents a representative example in which only \name produces a syntactically valid output.
In this example, the inference process of \baselinename{IG-CD} stalls with a single remaining \texttt{[MASK]} token. Theoretically, completing the output at this point requires multiple tokens (such as a variable name, a closing parenthesis, and a semicolon). However, since only one masked position remains, it becomes impossible for the model to generate a grammatically correct completion. We attribute this failure to the unreliable constraints imposed by IG-CD, which allow the model to enter a state where valid continuation is no longer possible.
In contrast, \baselinename{FS-CD} suffers from severe repetition. As shown in Figure~\ref{fig:rq1}, the model repeatedly generates the same \texttt{if} statement until the output reaches the maximum generation length, resulting in a forced termination. We attribute this issue to FS-CD's strict enforcement of left-to-right generation, which disrupts the intrinsic non-autoregressive generation mechanism of dLLMs and degrades the quality of the predicted token distributions.

\cl{Since dLLMs exhibit randomness during inference, we additionally conduct statistical tests on the results of RQ1. Specifically, we repeat each experiment multiple times and examine the variance of all reported metrics. The results show that the maximum variance is only 1.52\%, indicating that the randomness has a negligible impact on our conclusions.}

\begin{boxK}
\textbf{Answer to RQ1:}
\name effectively enforces CFG constraints on generated outputs and outperforms all baselines across most benchmarks and models. Compared to existing approaches, \name ensures the validity of every newly generated token throughout the inference process.
\end{boxK}

\subsection{RQ2: How Well Does \name Improve the Functional Correctness of dLLM Outputs?}
\label{sec:rq2}

\textbf{Motivation.}
Although the primary objective of CFG-constrained decoding is to guarantee the syntactic correctness of generated outputs, we are also interested in whether \name can improve the functional correctness of generated outputs. In this RQ, we evaluate the impact of \name on the functional correctness of dLLM outputs.

\begin{cp}
\vspace{3pt}
\noindent \textbf{Setting.}
We apply the baselines and \name to the four dLLMs described in Section~\ref{sec:models}. We then measure the performance of different approaches on \dataname{JSON-Bench}, \dataname{CPP-Bench}, \dataname{JAVA-Bench}, and \dataname{GO-Bench}. 
As discussed in Section~\ref{sec:rq1}, \baselinename{IG-CD} is difficult to evaluate on \dataname{GO-Bench} and \dataname{JAVA-Bench}, and is therefore excluded from these two benchmarks.
We use \textit{functional@k} as the evaluation metric with $k \in \{1, 5, 10\}$, as defined in Section~\ref{sec:metric}.
We exclude \dataname{SMILES-Bench} because all models yielded a 0\% functional correctness rate on this benchmark, making meaningful comparison infeasible.
\end{cp}

\begin{table}[t]
\cc
\centering
\small
\vspace{2pt}
\renewcommand{\arraystretch}{0.83}
\setlength{\tabcolsep}{0.5pt}
\caption{\cc
\textit{functional@k} (\%) on \dataname{JSON-Bench}, \dataname{CPP-Bench}, \dataname{JAVA-Bench}, and \dataname{GO-Bench}, where $k\in\{1,5,10\}$. 
The \textbf{best result} for each model, benchmark, and $k$ is highlighted in bold.
Absolute differences (in percentage points) are computed with respect to \baselinename{NO-CD}.}
\label{tab:rq2}
\vspace{-0.08in}
\scalebox{0.97}{
\begin{tabular}{c ccc ccc ccc ccc}
\toprule
\multirow{2}{*}{\textbf{Method}} &
\multicolumn{3}{c}{\textbf{\dataname{JSON-Bench}}} &
\multicolumn{3}{c}{\textbf{\dataname{CPP-Bench}}} &
\multicolumn{3}{c}{\textbf{\dataname{JAVA-Bench}}} & 
\multicolumn{3}{c}{\textbf{\dataname{GO-Bench}}}\\
\cmidrule(lr){2-4}\cmidrule(lr){5-7}\cmidrule(lr){8-10}\cmidrule(lr){11-13}
& \textbf{k=1} & \textbf{k=5} & \textbf{k=10}
& \textbf{k=1} & \textbf{k=5} & \textbf{k=10}
& \textbf{k=1} & \textbf{k=5} & \textbf{k=10}
& \textbf{k=1} & \textbf{k=5} & \textbf{k=10} \\
\midrule

\multicolumn{13}{c}{\textbf{\llmname{LLaDA-8B}}} \\
\arrayrulecolor{gray}\midrule\arrayrulecolor{black}
\baselinename{NO-CD}
& 46.9\same{+0.0} & 49.2\same{+0.0} & 49.6\same{+0.0}
& 16.3\same{+0.0} & 23.1\same{+0.0} & 24.9\same{+0.0} 
& 14.0\same{+0.0} & 18.3\same{+0.0} & 21.3\same{+0.0}
& 12.8\same{+0.0} & 18.3\same{+0.0} & 19.5\same{+0.0} \\
\baselinename{FS-CD}
& 52.9\up{+6.0} & 53.3\up{+4.1} & 53.5\up{+3.9}
& 15.7\down{-0.6} & 16.0\down{-7.1} & 16.2\down{-8.7}
& 15.1\up{+1.1} & 15.2\down{-3.1} & 15.2\down{-6.1}
& 11.6\down{-1.2} & 13.0\down{-5.3} & 13.4\down{-7.9} \\
\baselinename{IG-CD}
& 52.1\up{+5.2} & 53.3\up{+4.1} & 53.7\up{+4.1}
& 17.8\up{+1.5} & 23.7\up{+0.6} & 25.6\up{+0.7} 
& -- & -- & --
& -- & -- & -- \\
\rowcolor[HTML]{F3F3F3}
\name
& \textbf{55.0\up{+8.1}} & \textbf{56.3\up{+7.1}} & \textbf{56.4\up{+6.8}}
& \textbf{19.5\up{+3.2}} & \textbf{25.9\up{+2.8}} & \textbf{26.7\up{+1.8}} 
& \textbf{18.3\up{+4.3}} & \textbf{22.0\up{+3.7}} & \textbf{24.4\up{+3.1}} 
& \textbf{17.7\up{+4.9}} & \textbf{21.3\up{+3.0}} & \textbf{22.6\up{+3.1}} \\
\midrule

\multicolumn{13}{c}{\textbf{\llmname{LLaDA-1.5}}} \\
\arrayrulecolor{gray}\midrule\arrayrulecolor{black}
\baselinename{NO-CD}
& 45.5\same{+0.0} & 46.5\same{+0.0} & 46.7\same{+0.0}
& 17.2\same{+0.0} & 22.0\same{+0.0} & 25.9\same{+0.0} 
& 18.3\same{+0.0} & 21.3\same{+0.0} & 22.6\same{+0.0}
& 15.2\same{+0.0} & 16.5\same{+0.0} & 17.7\same{+0.0} \\
\baselinename{FS-CD}
& 53.1\up{+7.6} & 53.6\up{+7.1} & 53.6\up{+6.9}
& 15.9\down{-1.3} & 16.5\down{-5.5} & 16.5\down{-9.4} 
& 15.2\down{-3.1} & 15.2\down{-6.1} & 15.2\down{-7.4}
& 15.7\up{+0.5} & 15.9\down{-0.6} & 15.9\down{-1.8} \\
\baselinename{IG-CD}
& 50.6\up{+5.1} & 52.2\up{+5.8} & 52.5\up{+5.8}
& 18.3\up{+1.1} & 23.1\up{+1.1} & 27.1\up{+1.2} 
& -- & -- & --
& -- & -- & -- \\
\rowcolor[HTML]{F3F3F3}
\name
& \textbf{54.1\up{+8.6}} & \textbf{54.7\up{+8.2}} & \textbf{54.7\up{+8.0}}
& \textbf{19.1\up{+1.9}} & \textbf{24.9\up{+2.9}} & \textbf{28.8\up{+2.9}} 
& \textbf{20.7\up{+2.4}} & \textbf{23.2\up{+1.9}} & \textbf{24.4\up{+1.8}}
& \textbf{18.3\up{+3.1}} & \textbf{19.1\up{+2.6}} & \textbf{19.5\up{+1.8}} \\
\midrule

\multicolumn{13}{c}{\textbf{\llmname{Dream-7B}}} \\
\arrayrulecolor{gray}\midrule\arrayrulecolor{black}
\baselinename{NO-CD}
& 49.8\same{+0.0} & 50.7\same{+0.0} & 50.7\same{+0.0}
& 25.6\same{+0.0} & 26.8\same{+0.0} & 28.1\same{+0.0} 
& 24.4\same{+0.0} & 25.6\same{+0.0} & 25.6\same{+0.0}
& 24.4\same{+0.0} & 26.8\same{+0.0} & 26.8\same{+0.0} \\
\baselinename{FS-CD}
& 52.9\up{+3.1} & 53.2\up{+2.5} & 53.2\up{+2.5}
& 26.5\up{+0.9} & 27.1\up{+0.3} & 27.5\down{-0.6} 
& 25.0\up{+0.6} & 25.0\down{-0.6} & 25.0\down{-0.6}
& 27.9\up{+3.5} & 28.0\up{+1.2} & 28.0\up{+1.2} \\
\baselinename{IG-CD}
& 51.5\up{+1.7} & 53.5\up{+2.8} & 53.9\up{+3.2}
& 29.1\up{+3.5} & 30.8\up{+4.0} & 31.1\up{+3.0} 
& -- & -- & --
& -- & -- & -- \\
\rowcolor[HTML]{F3F3F3}
\name
& \textbf{54.4\up{+4.6}} & \textbf{55.1\up{+4.4}} & \textbf{55.5\up{+4.8}}
& \textbf{33.5\up{+7.9}} & \textbf{39.5\up{+12.7}} & \textbf{41.4\up{+13.3}} 
& \textbf{28.1\up{+3.7}} & \textbf{29.3\up{+3.7}} & \textbf{30.5\up{+4.9}}
& \textbf{30.8\up{+6.4}} & \textbf{32.3\up{+5.5}} & \textbf{33.1\up{+6.3}} \\
\midrule

\multicolumn{13}{c}{\textbf{\llmname{DiffuC-7B}}} \\
\arrayrulecolor{gray}\midrule\arrayrulecolor{black}
\baselinename{NO-CD}
& 52.0\same{+0.0} & 53.3\same{+0.0} & 53.7\same{+0.0}
& 37.2\same{+0.0} & 37.8\same{+0.0} & 37.8\same{+0.0} 
& 29.3\same{+0.0} & 30.5\same{+0.0} & 30.5\same{+0.0} 
& 22.0\same{+0.0} & 22.6\same{+0.0} & 23.3\same{+0.0} \\
\baselinename{FS-CD}
& 53.7\up{+1.7} & 54.1\up{+0.8} & 54.1\up{+0.4}
& 35.2\down{-2.0} & 35.8\down{-2.0} & 35.9\down{-1.9} 
& 29.3\same{+0.0} & 29.7\down{-0.8} & 29.9\down{-0.6}
& 23.2\up{+1.2} & 23.2\up{+0.6} & 23.2\down{-0.1} \\
\baselinename{IG-CD}
& 53.0\up{+1.0} & 55.3\up{+2.0} & 55.3\up{+1.6}
& \textbf{38.1\up{+0.9}} & 38.3\up{+0.5} & \textbf{38.5\up{+0.7}} 
& -- & -- & --
& -- & -- & -- \\
\rowcolor[HTML]{F3F3F3}
\name
& \textbf{55.2\up{+3.2}} & \textbf{56.6\up{+3.3}} & \textbf{56.7\up{+3.0}}
& \textbf{38.1\up{+0.9}} & \textbf{38.5\up{+0.7}} & \textbf{38.5\up{+0.7}} 
& \textbf{32.2\up{+2.9}} & \textbf{32.9\up{+2.4}} & \textbf{32.9\up{+2.4}}
& \textbf{24.4\up{+2.2}} & \textbf{25.8\up{+3.2}} & \textbf{26.2\up{+2.9}} \\
\bottomrule
\end{tabular}
}
\vspace{-0.1in}
\end{table}

\vspace{3pt}
\noindent \textbf{Results.}
The \textit{functional@k} of different approaches is reported in Table~\ref{tab:rq2}.

\textbf{\name consistently improves the functional correctness of dLLMs across four models and four benchmarks.} Compared to the unconstrained baseline, \name achieves higher \textit{functional@k} in all settings. In contrast, other constrained decoding baselines sometimes degrade model performance. For example, on \dataname{CPP-Bench} with \llmname{LLaDA-8B}, \baselinename{FS-CD} reduces \textit{functional@1} by 0.6\%, whereas \name increases it by 3.2\%.
\cl{A similar pattern can also be observed on \dataname{JAVA-Bench}, where, in terms of \textit{functional@10}, \baselinename{FS-CD} decreases the score by 6.1\%, while \name improves it by 3.1\%.}

\textbf{\name outperforms other CFG-constrained decoding approaches in improving functional correctness.} The gains achieved by \name are larger than those of the baseline methods for most settings. For example, on \llmname{Dream-7B} with \dataname{CPP-Bench}, \name improves \textit{functional@1}, \textit{functional@5}, and \textit{functional@10} from 25.6\%, 26.8\%, and 28.1\% under the unconstrained setting to 33.5\%, 39.5\%, and 41.4\%, respectively. In contrast, the best baseline reaches only 29.1\%, 30.8\%, and 31.1\%.

\textbf{\name yields larger functional correctness gains on tasks with limited reasoning requirements.}
\dataname{JSON-Bench} focuses on structured information extraction and typically does not require complex logical reasoning, whereas \dataname{CPP-Bench} involves code generation and is more dependent on reasoning ability. We observe that, compared to \baselinename{NO-CD}, \name achieves larger improvements in \textit{functional@k} on \dataname{JSON-Bench} than on \dataname{CPP-Bench}. For example, with \llmname{LLaDA-8B}, \name improves functional correctness by 8.1\% on \dataname{JSON-Bench}, whereas the improvement is 3.2\% on \dataname{CPP-Bench}. This behavior is expected because, in tasks with limited reasoning requirements, functional correctness depends more strongly on syntactic validity, allowing the syntactic improvements provided by \name to translate more effectively into functional gains.

\cl{Similar to RQ1, we repeat each experiment multiple times and examine the variance of all reported metrics. The results show that the maximum standard deviation is only 0.98\%.}

\begin{boxK}
\textbf{Answer to RQ2:}
Compared to baselines, \name improves the functional correctness of dLLMs more consistently and significantly. Furthermore, this improvement is particularly evident in tasks with lower logical reasoning requirements.
\end{boxK}

\begin{cp}
\subsection{RQ3: What Runtime Overhead Does \name Introduce?}
\end{cp}
\label{sec:rq3}

\begin{cp}
\noindent
\textbf{Motivation.}
Beyond the quality of generated outputs, runtime efficiency is also a critical factor for the practical adoption of constrained decoding approaches. If a constrained decoding method introduces excessive inference overhead, its practical utility would be substantially limited~\cite{beurer2024guiding}. Therefore, in this RQ, we examine the runtime overhead introduced by \name.

\vspace{3pt}
\noindent \textbf{Setting.}
We apply the baselines and \name to the four models described in Section~\ref{sec:models} and evaluate them on the five benchmarks introduced in Section~\ref{sec:benchmark}. We use average inference time as the metric to measure the runtime overhead of different approaches.
As discussed in Section~\ref{sec:rq1}, \baselinename{IG-CD} is difficult to evaluate on \dataname{GO-Bench} and \dataname{JAVA-Bench}, and is therefore excluded from these two benchmarks.
\end{cp}

\begin{table}[t]
\vspace{2pt}
\cc
\centering
\small
\renewcommand{\arraystretch}{1}
\setlength{\tabcolsep}{0.8pt}
\caption{
\cc
Average inference time (s) on \dataname{JSON-Bench}, \dataname{CPP-Bench}, \dataname{JAVA-Bench}, \dataname{GO-Bench}, and \dataname{SMILES-Bench}.
The best result for each model and benchmark is highlighted in bold. 
The last row reports the relative inference time normalized by \baselinename{NO-CD}.}
\label{tab:rq3}
\vspace{-0.1in}
\scalebox{0.77}{
\begin{tabular}{c|cccc|cccc|ccc|ccc|cccc}
\toprule
\multirow{2}{*}{\textbf{Model}} 
& \multicolumn{4}{c|}{\textbf{\dataname{JSON-Bench}}} 
& \multicolumn{4}{c|}{\textbf{\dataname{CPP-Bench}}} 
& \multicolumn{3}{c|}{\textbf{\dataname{JAVA-Bench}}} 
& \multicolumn{3}{c|}{\textbf{\dataname{GO-Bench}}} 
& \multicolumn{4}{c}{\textbf{\dataname{SMILES-Bench}}} \\
\cmidrule(lr){2-5}
\cmidrule(lr){6-9}
\cmidrule(lr){10-12}
\cmidrule(lr){13-15}
\cmidrule(lr){16-19}
& \baselinename{NO-CD} & \baselinename{FS-CD} & \baselinename{IG-CD} & \name
& \baselinename{NO-CD} & \baselinename{FS-CD} & \baselinename{IG-CD} & \name
& \baselinename{NO-CD} & \baselinename{FS-CD} & \name
& \baselinename{NO-CD} & \baselinename{FS-CD} & \name
& \baselinename{NO-CD} & \baselinename{FS-CD} & \baselinename{IG-CD} & \name \\
\midrule
\llmname{LLaDA-8B}
& {9.48} & 13.29 & 10.34 & \textbf{8.22} 
& \textbf{8.55} & 16.68 & 9.66 & {9.30} 
& \textbf{8.03} & 8.74 & {8.56}
& \textbf{8.09} & 10.19 & {8.32}
& 8.49 & {7.68} & 8.76 & \textbf{3.02} \\
\llmname{LLaDA-1.5}
& \textbf{9.71} & 14.74 & 10.82 & {9.91} 
& \textbf{8.56} & {11.88} & 12.17 & 12.02
& \textbf{8.16} & 10.29 & {9.22}
& \textbf{8.23} & {8.28} & 9.08
& 8.77 & \textbf{2.16} & 8.21 & {2.97}  \\
\llmname{Dream-7B}
& {8.62} & 15.74 & \textbf{8.25} & 9.73 
& 7.46 & \textbf{7.17} & 7.95 & {7.26}  
& {7.37} & \textbf{6.30} & 6.41
& \textbf{7.05} & 10.01 & {7.67}
& {7.07} & 8.99 & 19.64 & \textbf{2.24}  \\
\llmname{DiffuC-7B}
& \textbf{8.15} & 15.39 & 9.78 & {9.22} 
& \textbf{7.45} & 14.78 & {8.47} & 12.27
& {7.38} & \textbf{6.81} & 7.52
& \textbf{7.03} & {7.11} & 7.40
& 6.86 & 11.15 & {6.74} & \textbf{4.41}  \\
\midrule
Average      
& \textbf{8.99} & 14.79 & 9.80 & {9.27} 
& \textbf{8.01} & 12.63 & {9.56} & 10.21 
& \textbf{7.74} & 8.04 & {7.93}
& \textbf{7.60} & 8.90 & {8.12}
& 7.80 & {7.50} & 10.84 & \textbf{3.16} \\
Ratio    
& $\times$1.00 
& $\times$1.65 & $\times$1.09 & $\times$1.03
& $\times$1.00 
& $\times$1.58 & $\times$1.19 & $\times$1.27
& $\times$1.00 
& $\times$1.04 & $\times$1.02
& $\times$1.00 
& $\times$1.17 & $\times$1.07
& $\times$1.00 
& $\times$0.96 & $\times$1.39 & $\times$0.41 \\
\bottomrule
\end{tabular}
}
\end{table}

\vspace{3pt}
\noindent \textbf{Results.}
The average inference times of different approaches are reported in Table~\ref{tab:rq3}. Lower values indicate better efficiency.

\textbf{\name introduces negligible runtime overhead across most benchmarks and models.}
For example, on \llmname{LLaDA-1.5} evaluated on \dataname{JSON-Bench}, applying \name increases the average inference time from 9.71 seconds to 9.91 seconds, corresponding to an increase of only 2\%. For \llmname{LLaDA-8B} on \dataname{CPP-Bench}, where the grammar of C++ is considerably more complex than that of JSON or SMILES, the inference time increases modestly from 8.55 seconds to 9.30 seconds, corresponding to a 9\% increase.
Interestingly, we observe that in several scenarios, the average inference time of \name is even lower than that of the unconstrained baseline. For example, with \llmname{DiffuC-7B} on \dataname{SMILES-Bench}, the average inference time for \name is 4.41 seconds, compared to 6.86 seconds for the unconstrained setting. We attribute this phenomenon to the fact that, without constraints, the model often generates syntactically invalid outputs (such as extraneous natural language explanations), resulting in more generated tokens and consequently prolonging the inference process.

\vspace{1pt}
\begin{boxK}
\textbf{Answer to RQ3:}
\name introduces negligible runtime overhead in most settings, allowing it to improve the reliability of structured generation in dLLMs with minimal additional cost.
\end{boxK}

\subsection{RQ4: How Does Each Component of \name Contribute to the Overall Performance?}
\label{sec:rq4}

\begin{table*}[!t]
\centering
\small
\renewcommand{\arraystretch}{0.88}
\setlength{\tabcolsep}{2.2pt}
\caption{Ablation study of \name. 
Verification and Recovery denote Lookahead-Based Verification and Cache-Enhanced Recovery, respectively. 
Syn and Func represent \textit{syntactic@1} and \textit{functional@1}, respectively, while Time denotes the average inference time.}
\label{tab:rq4}
\vspace{-0.1in}
\scalebox{1}{
\begin{tabular}{c|cc|ccc|ccc}
\toprule
\multirow{2}{*}{\textbf{Model}} 
& \multirow{2}{*}{\textbf{Verification}} & \multirow{2}{*}{\textbf{Recovery}}
& \multicolumn{3}{c|}{\textbf{\dataname{JSON-Bench}}}
& \multicolumn{3}{c}{\textbf{\dataname{CPP-Bench}}} \\
\cmidrule(lr){4-6}
\cmidrule(lr){7-9}
& & 
& Syn.\ (\%) & Func.\ (\%) & Time (s)
& Syn.\ (\%) & Func.\ (\%) & Time (s) \\
\midrule

\multirow{3}{*}{\llmname{LLaDA-8B}}
& \ding{55} & \ding{55}
& 85.4 \textcolor[HTML]{909090}{(+0.0)} & 46.9 \textcolor[HTML]{909090}{(+0.0)} & 9.48
& 74.2 \textcolor[HTML]{909090}{(+0.0)} & 16.3 \textcolor[HTML]{909090}{(+0.0)} & 8.55 \\
& \ding{51} & \ding{55}
& 95.8 \textcolor[HTML]{8B0000}{(+10.4)} & 53.7 \textcolor[HTML]{8B0000}{(+6.8)} & 8.09
& 86.6 \textcolor[HTML]{8B0000}{(+12.4)} & 18.9 \textcolor[HTML]{8B0000}{(+2.6)} & 8.71 \\
& \ding{51} & \ding{51}
& 99.5 \textcolor[HTML]{8B0000}{(+14.1)} & 55.0 \textcolor[HTML]{8B0000}{(+8.1)} & 8.22
& 90.9 \textcolor[HTML]{8B0000}{(+16.7)} & 19.5 \textcolor[HTML]{8B0000}{(+3.2)} & 9.30 \\

\midrule

\multirow{3}{*}{\llmname{Dream-7B}}
& \ding{55} & \ding{55}
& 86.8 (+0.0) & 49.8 (+0.0) & 8.62
& 76.2 (+0.0) & 25.6 (+0.0) & 7.46 \\
& \ding{51} & \ding{55}
& 94.8 \textcolor[HTML]{8B0000}{(+8.0)} & 51.8 \textcolor[HTML]{8B0000}{(+2.0)} & 9.17
& 92.7 \textcolor[HTML]{8B0000}{(+16.5)} & 32.3 \textcolor[HTML]{8B0000}{(+6.7)} & 7.69 \\
& \ding{51} & \ding{51}
& 99.0 \textcolor[HTML]{8B0000}{(+12.2)} & 54.4 \textcolor[HTML]{8B0000}{(+4.6)} & 9.73
& 96.0 \textcolor[HTML]{8B0000}{(+19.8)} & 33.5 \textcolor[HTML]{8B0000}{(+7.9)} & 7.26 \\

\bottomrule
\end{tabular}
}
\vspace{-0.1in}
\end{table*}

\textbf{Motivation.}
\name consists of two components, namely Lookahead-Based Verification and Cache-Enhanced Recovery. In this RQ, we evaluate the contribution of each component through ablation experiments.

\vspace{3pt}
\noindent \textbf{Setting.}
\cl{Due to the high computational cost, we select two representative models and two representative benchmarks for ablation experiments. Specifically, we conduct experiments on \llmname{LLaDA-8B} and \llmname{Dream-7B} using \dataname{JSON-Bench} and \dataname{CPP-Bench}.}
All experiments are evaluated using the three metrics defined in Section~\ref{sec:metric}.  
Since Cache-Enhanced Recovery depends on Lookahead-Based Verification, it is not possible to remove Lookahead-Based Verification while retaining Cache-Enhanced Recovery. Therefore, we consider three experimental settings:  
\ding{182} retaining both Lookahead-Based Verification and Cache-Enhanced Recovery, corresponding to \name;  
\ding{183} retaining Lookahead-Based Verification while removing Cache-Enhanced Recovery, where the model switches to unconstrained generation once the triggering condition for Cache-Enhanced Recovery is met;  
\ding{184} removing both Lookahead-Based Verification and Cache-Enhanced Recovery, which corresponds to \baselinename{NO-CD}.

\vspace{3pt}
\noindent \textbf{Results.}
The results of the ablation experiments are reported in Table~\ref{tab:rq4}.

\textbf{Lookahead-Based Verification alone significantly improves both syntactic and functional correctness.}
For example, on \llmname{LLaDA-8B} evaluated on \dataname{JSON-Bench}, applying Lookahead-Based Verification increases the \textit{syntactic@1} by 10.4\% and the \textit{functional@1} by 6.8\%. On \llmname{Dream-7B} evaluated on \dataname{CPP-Bench}, applying Lookahead-Based Verification increases the \textit{syntactic@1} by 16.5\% and the \textit{functional@1} by 6.7\%. 

\textbf{Cache-Enhanced Recovery further improves the performance of \name with negligible additional runtime overhead.}
Across all evaluated settings, Cache-Enhanced Recovery consistently improves both syntactic and functional correctness compared to using Lookahead-Based Verification alone, while having a negligible impact on average inference time.
For example, on \llmname{Dream-7B} evaluated on \dataname{JSON-Bench}, it increases the \textit{syntactic@1} to 99.0\% and the \textit{functional@1} to 54.4\%, with an added runtime cost of less than 0.6 seconds.

\begin{boxK}
\textbf{Answer to RQ4:}
Both Lookahead-Based Verification and Cache-Enhanced Recovery contribute effectively to the performance of \name by improving the syntactic and functional correctness of dLLM outputs.
\end{boxK}

\subsection{RQ5: How Do the Hyperparameters Affect the Performance of \name?}
\label{sec:rq5}

\begin{figure}[t]
\vspace{1pt}
\centering

\begin{subfigure}[t]{0.48\textwidth}
    \centering
    \includegraphics[width=\linewidth]{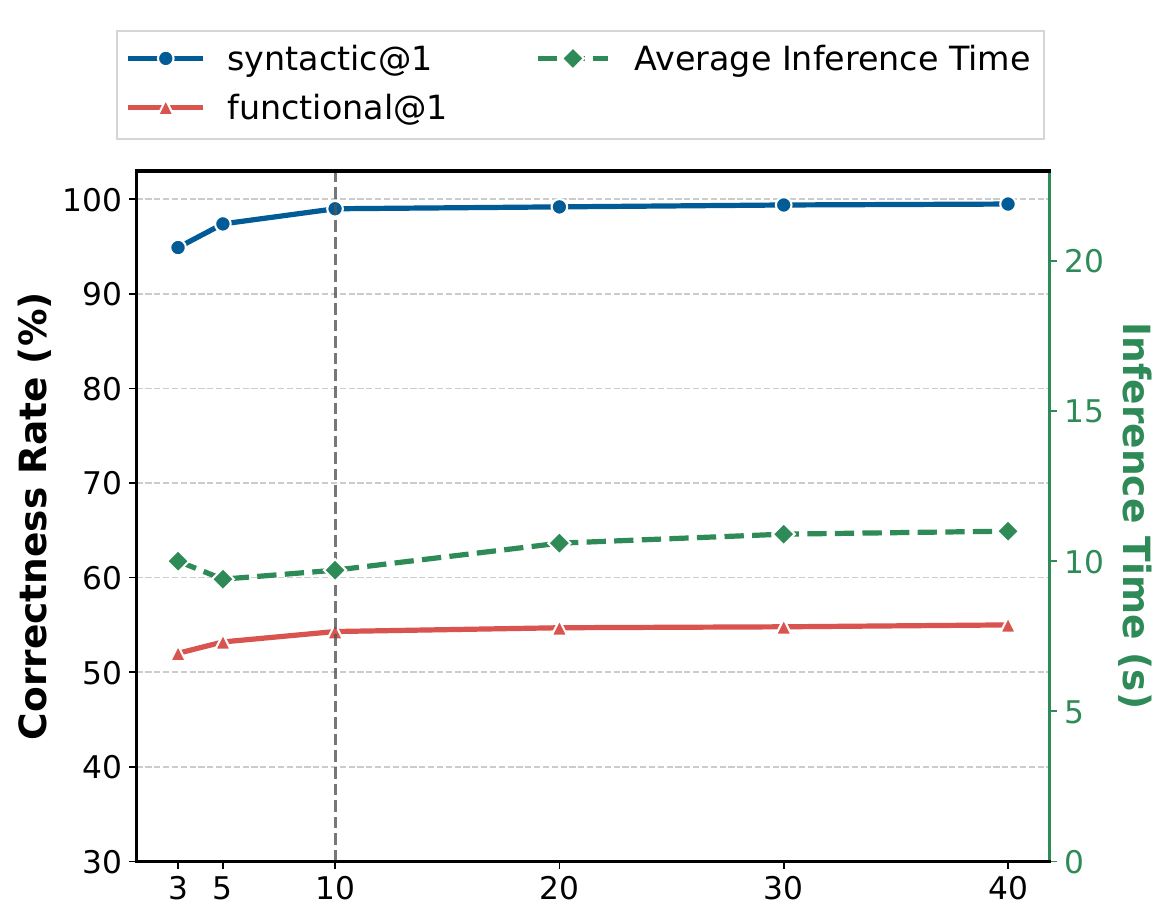}%
    \vspace{-0.08in}
    \caption{Lookahead size $N$}
    \label{fig:rq5-N}
\end{subfigure}
\hfill
\begin{subfigure}[t]{0.48\textwidth}
    \centering
    \includegraphics[width=\linewidth]{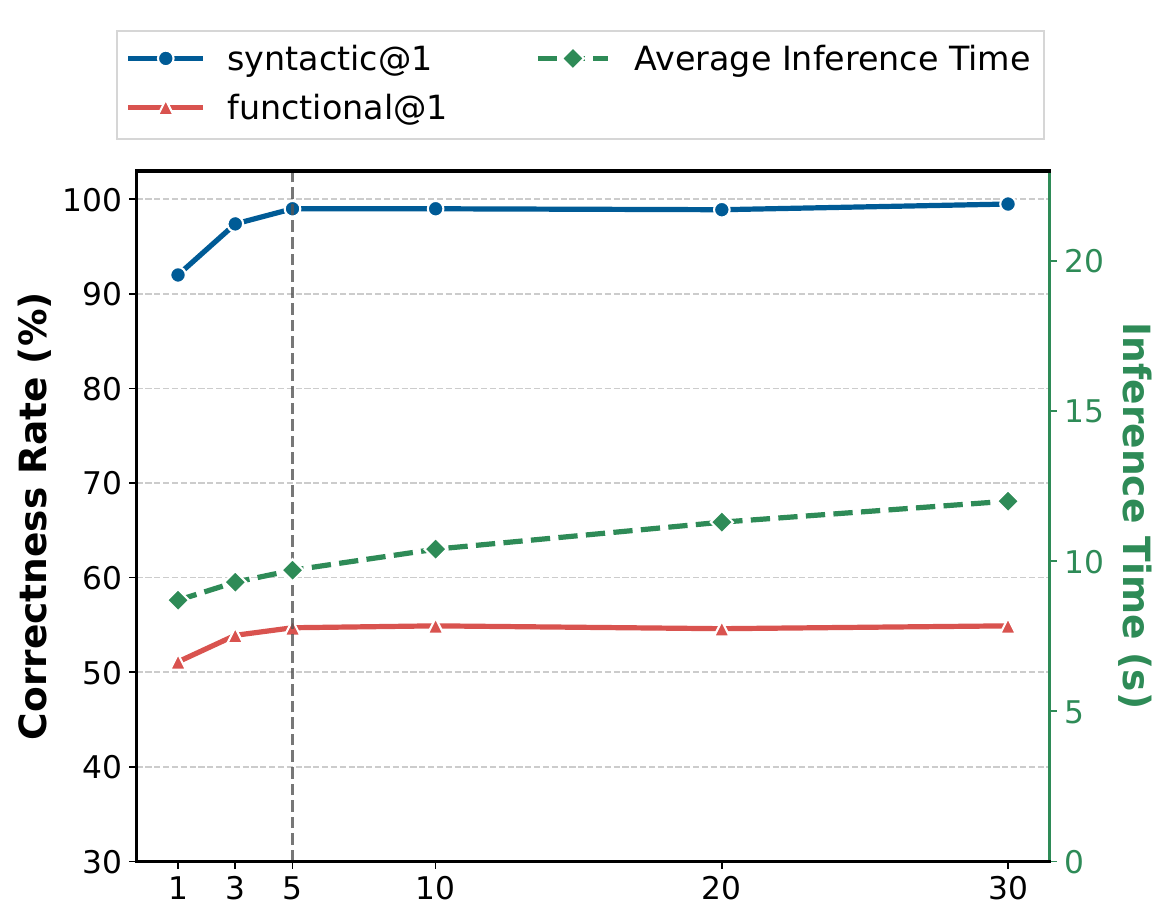}%
    \vspace{-0.08in}
    \caption{Proposal budget $\tau$}
    \label{fig:rq5-tau}
\end{subfigure}
\vspace{-0.11in}
\caption{The performance of \name with different values of the lookahead size $N$ and proposal budget $\tau$. We use the gray dashed line to represent the employed hyperparameters.}
\label{fig:rq5}
\vspace{-0.1in}
\end{figure}

\begin{table}[t]
\vspace{1pt}
\centering
\small
\renewcommand{\arraystretch}{1.0}
\setlength{\tabcolsep}{2.1pt}
\caption{Performance of \name with varying values of the hyperparameter $T$. 
The gray background indicates the hyperparameter setting used in our experiments.}
\label{tab:rq5}
\vspace{-0.1in}
\scalebox{0.9}{
\begin{tabular}{c|
    c c c >{\columncolor[HTML]{F3F3F3}}c c |
    c c c >{\columncolor[HTML]{F3F3F3}}c c |
    c c c >{\columncolor[HTML]{F3F3F3}}c c}
\toprule
\multirow{2}{*}{\textbf{Method}} 
& \multicolumn{5}{c|}{\textbf{\textit{syntactic@1} (\%)}} 
& \multicolumn{5}{c|}{\textbf{\textit{functional@1} (\%)}} 
& \multicolumn{5}{c}{\textbf{Average Inference Time (s)}} \\
\cmidrule(lr){2-6}
\cmidrule(lr){7-11}
\cmidrule(lr){12-16}
& $T{=}16$ & $T{=}32$ & $T{=}64$ & $T{=}128$ & $T{=}256$
& $T{=}16$ & $T{=}32$ & $T{=}64$ & $T{=}128$ & $T{=}256$
& $T{=}16$ & $T{=}32$ & $T{=}64$ & $T{=}128$ & $T{=}256$ \\
\midrule
\baselinename{NO-CD}
& 7.8 & 21.0 & 63.6 & 86.8 & 97.4
& 1.1 & 7.4 & 40.4 & 49.8 & 54.8
& 1.98 & 2.92 & 4.56 & 8.62 & 15.88 \\
\name
& 76.3 & 91.7 & 97.2 & 99.0 & 100.0
& 20.4 & 32.2 & 46.6 & 54.4 & 54.8
& 3.21 & 3.76 & 5.48 & 9.73 & 16.02 \\
\bottomrule
\end{tabular}
}
\vspace{-0.12in}
\end{table}

\textbf{Motivation.}
\name involves several hyperparameters that may affect its performance. In this RQ, we conduct a sensitivity analysis to examine their impact. Specifically, we consider two categories of hyperparameters. \ding{192} The first category consists of hyperparameters introduced by \name, including the lookahead size $N$, defined in Section~\ref{sec:check}, and the attempt budget $\tau$, defined in Section~\ref{sec:recovery}. \ding{193} The second category includes key hyperparameters inherent to dLLMs, for which we select the number of denoising steps $T$.

\vspace{3pt}
\noindent \textbf{Setting.}
\cl{Due to the high computational cost,} we conduct experiments on \llmname{Dream-7B} using \dataname{JSON-Bench}. All experiments are evaluated using the three metrics defined in Section~\ref{sec:metric}. For the lookahead size $N$, we consider values in $\{3, 5, 10, 20, 30, 40\}$. For the attempt budget $\tau$, we consider values in $\{1, 3, 5, 10, 20, 30\}$. For the number of denoising steps $T$, we consider values in $\{16, 32, 64, 128, 256\}$.

\vspace{3pt}
\noindent \textbf{Results.}

\textit{The results for the hyperparameters introduced by \name, namely the lookahead size $N$ and the attempt budget $\tau$, are shown in Figure~\ref{fig:rq5}.}
\ding{182} \textbf{Effect of the lookahead size $N$.}
We observe that increasing $N$ generally leads to improvements in both \textit{syntactic@1} and \textit{functional@1}. This trend can be attributed to the fact that a larger $N$ increases the likelihood of identifying at least one valid \textit{complete prefix} during Lookahead-Based Verification. 
Interestingly, the average inference time first decreases and then increases as $N$ grows. This behavior reflects two competing effects. On the one hand, a larger $N$ increases the number of \textit{complete prefixes} evaluated during each validation step, which increases inference time. On the other hand, a larger $N$ increases the probability that a valid proposed token is accepted, thereby reducing the number of re-proposal attempts and shortening the overall inference time.
\ding{183} \textbf{Effect of the proposal budget $\tau$.}
As $\tau$ increases, all three metrics exhibit an upward trend. A larger attempt budget allows the model more opportunities to propose high-confidence tokens, which improves both syntactic and functional correctness. At the same time, increasing $\tau$ leads to more re-proposal, which increases inference time.
\ding{184} \textbf{Overall robustness to hyperparameter variations.}
Overall, \name demonstrates strong robustness to adjustments in the newly introduced hyperparameters. Although varying lookahead size $N$ and $\tau$ leads to observable performance trends, the magnitude of these changes remains limited. For example, when $N$ increases from 3 to 40, the average inference time remains stable at around 10 seconds. Similarly, when $\tau$ increases from 3 to 30, the \textit{syntactic@1} consistently remains above 97\%.
\textit{The results for the hyperparameters inherent to dLLMs, namely the denoising steps $T$, are shown in Table~\ref{tab:rq5}.}
\ding{182} \textbf{\name consistently improves syntactic and functional correctness across different values of $T$.}
For instance, when $T$ is set to 32, which in our setting corresponds to generating at least eight tokens per forward pass, \name increases the \textit{syntactic@1} from 21.0\% to 91.7\% and the \textit{functional@1} from 7.4\% to 32.2\%. Although the runtime overhead increases slightly when the number of denoising steps is small (\eg rising from 2.92s to 3.76s at $T=32$), we consider this cost acceptable given the substantial gains in correctness.
\ding{183} \textbf{The syntactic and functional correctness of dLLMs with \name increase as $T$ increases.}
For instance, as $T$ increases from 16 to 128, the \textit{syntactic@1} improves from 76.3\% to 99.0\%, and the \textit{functional@1} rises from 20.4\% to 54.4\%. 
We attribute this to the characteristics of dLLMs, where a higher number of denoising steps leads to better effectiveness but lower efficiency. The unconstrained baseline exhibits a similar pattern, further supporting this explanation.

\begin{boxK}
\textbf{Answer to RQ5:}
\name exhibits strong robustness to variations in key hyperparameters and consistently improves the quality of generated outputs across different settings.
\end{boxK}

\section{Discussion}
\label{sec:discussion}

\subsection{Understanding the Remaining Syntax Errors}
\label{sec:error}

Although our approach guarantees reliable grammar constraints by ensuring that no newly generated token renders the intermediate output impossible to complete into a valid sentence, occasional syntax errors still occur (Section~\ref{sec:rq1}). We identify the root cause of these remaining errors as the model's tendency to generate repetitive patterns, a phenomenon widely observed in previous studies~\cite{fu2021theoretical, holtzman2019curious, li2023repetition}. In such cases, the dLLM repeatedly generates specific content (\eg whitespace or newline characters) until it reaches the maximum generation length. To confirm this, we examined all invalid outputs from our experiments and found that the number of generated tokens exactly matched the maximum generation length in every case. Fortunately, the majority of approaches~\cite{dong2025rethinking, li2023repetition, xu2022learning} designed to resolve such repetition problems are orthogonal to our approach; therefore, they can be effectively combined with \name to further improve the reliability of dLLM generation.

\begin{cp}

\subsection{Comparison with AR LLMs}

\begin{table*}[!t]
\vspace{1pt}
\cc
\centering
\small
\renewcommand{\arraystretch}{0.83}
\setlength{\tabcolsep}{2.4pt}
\caption{\cc
Comparison with AR LLMs. 
The results are reported by \textit{syntactic@k} and \textit{functional@k} under $k\in\{1,5,10\}$ on \dataname{JSON-Bench} and \dataname{CPP-Bench}.}
\label{tab:ar}
\vspace{-7pt}
\scalebox{0.98}{
\begin{tabular}{c|c|ccc|ccc|ccc|ccc}
\toprule
\multirow{3}{*}{\textbf{Model}} 
& \multirow{3}{*}{\textbf{Method}}
& \multicolumn{6}{c|}{\textbf{\dataname{JSON-Bench}}}
& \multicolumn{6}{c}{\textbf{\dataname{CPP-Bench}}} \\
\cmidrule(lr){3-8}
\cmidrule(lr){9-14}
& 
& \multicolumn{3}{c|}{\textit{syntactic@k}}
& \multicolumn{3}{c|}{\textit{functional@k}}
& \multicolumn{3}{c|}{\textit{syntactic@k}}
& \multicolumn{3}{c}{\textit{functional@k}} \\
\cmidrule(lr){3-5}
\cmidrule(lr){6-8}
\cmidrule(lr){9-11}
\cmidrule(lr){12-14}
& 
& $k=1$ & $k=5$ & $k=10$
& $k=1$ & $k=5$ & $k=10$
& $k=1$ & $k=5$ & $k=10$
& $k=1$ & $k=5$ & $k=10$ \\
\midrule

\multirow{2}{*}{\llmname{LLaDA-8B}}
& w/o CD
& 91.2 & 91.2 & 91.2
& 50.0 & 50.7 & 51.1
& 98.2 & 98.8 & 98.8 
& 26.2 & 29.3 & 31.7 \\
& w/ CD
& 99.6 & 99.9 & \textbf{100.0}
& 54.0 & 54.8 & 55.2
& \textbf{99.4} & \textbf{100.0} & \textbf{100.0}
& 26.8 & 30.5 & 32.3 \\

\midrule

\multirow{2}{*}{\llmname{Dream-7B}}
& w/o CD
& 97.4 & 97.6 & 97.6
& 54.8 & 55.2 & 55.2
& 97.0 & 97.6 & 98.2
& 37.8 & 39.0 & 39.0 \\
& w/ CD
& \textbf{100.0} & \textbf{100.0} & \textbf{100.0}
& 54.8 & 56.3 & 57.0
& 98.8 & 99.4 & 99.4
& 39.0 & 39.9 & 40.2 \\

\midrule

\multirow{2}{*}{\llmname{LLaMA-8B}}
& w/o CD
& 92.3 & 93.4 & 94.9
& 55.2 & 57.4 & 57.7
& 93.3 & 96.3 & 97.0
& 37.8 & 44.5 & 47.6 \\
& w/ CD
& 99.6 & 99.6 & 99.6
& 56.3 & 57.4 & 57.7
& 94.2 & 96.6 & 97.0
& 38.4 & 45.1 & 48.2 \\

\midrule

\multirow{2}{*}{\llmname{Qwen-7B}}
& w/o CD
& 98.2 & 98.9 & 98.9
& 57.4 & \textbf{59.6} & \textbf{60.7}
& 93.3 & 97.6 & 97.6
& \textbf{60.4} & 68.9 & 69.5 \\
& w/ CD
& \textbf{100.0} & \textbf{100.0} & \textbf{100.0}
& \textbf{57.9} & \textbf{59.6} & \textbf{60.7}
& 98.6 & 98.8 & 98.8
& \textbf{60.4} & \textbf{69.5} & \textbf{70.7} \\

\bottomrule
\end{tabular}
}
\vspace{-0.14in}
\end{table*}

We further compare dLLMs with AR LLMs of similar scale on \dataname{JSON-Bench} and \dataname{CPP-Bench}. Specifically, we evaluate two representative dLLMs, \llmname{LLaDA-8B} and \llmname{Dream-7B}, and compare them with \llmname{Llama-3-8B-Instruct} and \llmname{Qwen2.5-7B-Instruct}, denoted as \llmname{LLaMA-8B} and \llmname{Qwen-7B}. For dLLMs, we compare unconstrained decoding with \name; for AR LLMs, we compare unconstrained decoding with \baselinename{LLGuidance}~\cite{llguidance}. For fairness, similar to AR LLMs, dLLMs are configured to decode one token per step. The results are shown in Table~\ref{tab:ar}.

\ding{182} \textbf{\name enables dLLMs to achieve strong syntactic correctness.}
For example, on \dataname{CPP-Bench}, \name improves the \textit{syntactic@5} of \llmname{LLaDA-8B} to 100.0\%, outperforming both constrained AR LLMs. On \dataname{JSON-Bench}, both dLLMs also achieve nearly perfect syntactic correctness with \name, confirming its effectiveness in enforcing CFG constraints.
\ding{183} \textbf{Current dLLMs still lag behind AR LLMs in functional correctness.}
Despite the syntactic gains, dLLMs remain generally weaker than strong AR LLMs in functional correctness. This mainly reflects the capability gap between current dLLMs and mature AR LLMs, suggesting that improving dLLM foundational capabilities is complementary to our work.
Moreover, although \llmname{Dream-7B} shows stronger functional correctness than \llmname{LLaDA-8B}, \name still consistently improves its correctness, further supporting that our approach is complementary to improvements in the foundational capabilities of dLLMs.

\subsection{Conditions for Efficient Lookahead}
\label{sec:efficient-lookahead}

The efficiency of \name relies on the empirical observation that only a small number of complete-prefix samples is usually sufficient for Lookahead-Based Verification (Section~\ref{sec:pre}). We acknowledge that this is not a formal guarantee: when \textit{the model is too weak, the grammar is too complex, or the problem is too difficult}, a larger lookahead size $N$ may be needed. Nevertheless, our experimental results show that this condition holds well across four popular dLLMs and five representative benchmarks, where \name consistently improves correctness with only modest runtime overhead. Importantly, this condition is used to justify the practicality of \name rather than the reliability of its design. The core contribution of \name is the reliability-oriented verification criterion, while the empirical small-$N$ observation explains why this criterion is affordable in practice.

\subsection{Understanding \name from a Correctness--Efficiency Perspective}

\begin{figure}[t]
\vspace{1pt}

\centering

\begin{subfigure}[t]{0.45\textwidth}
    \centering
    \begin{overpic}[width=\linewidth]{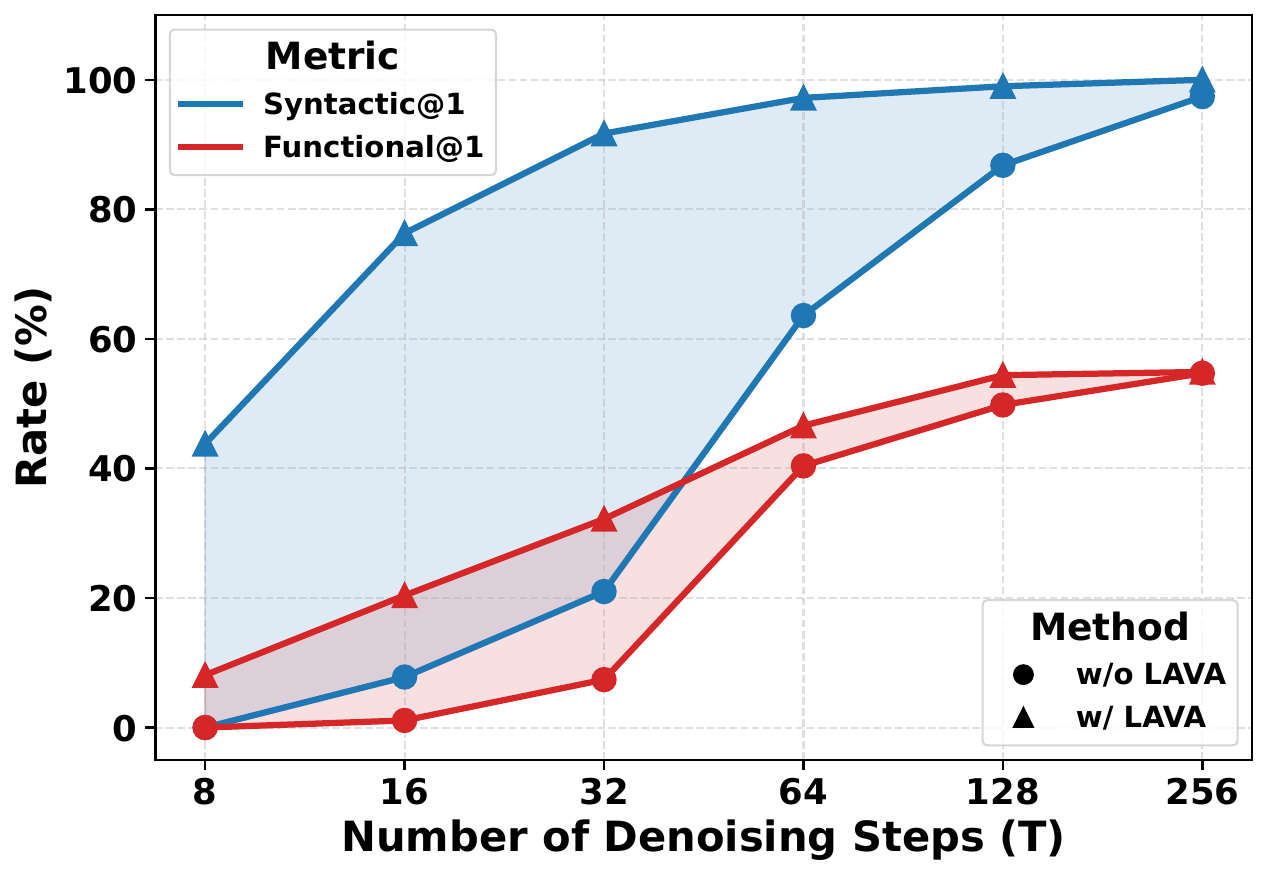}%
        \put(84.8,15.0){\begingroup\setlength{\fboxsep}{0pt}\colorbox{white}{\makebox[31pt][l]{\rule[-1.2pt]{0pt}{7.2pt}\fontsize{5.5pt}{6pt}\selectfont\sffamily\bfseries w/o LAVE}}\endgroup}
        \put(84.8,11.2){\begingroup\setlength{\fboxsep}{0pt}\colorbox{white}{\makebox[31pt][l]{\rule[-1.2pt]{0pt}{7.2pt}\fontsize{5.5pt}{6pt}\selectfont\sffamily\bfseries w/ LAVE}}\endgroup}
    \end{overpic}%
    \vspace{-0.03in}
    \caption{\cc
    \dataname{JSON-Bench}}
    \label{fig:trade-off-json}
\end{subfigure}
\hfill
\begin{subfigure}[t]{0.45\textwidth}
    \centering
    \begin{overpic}[width=\linewidth]{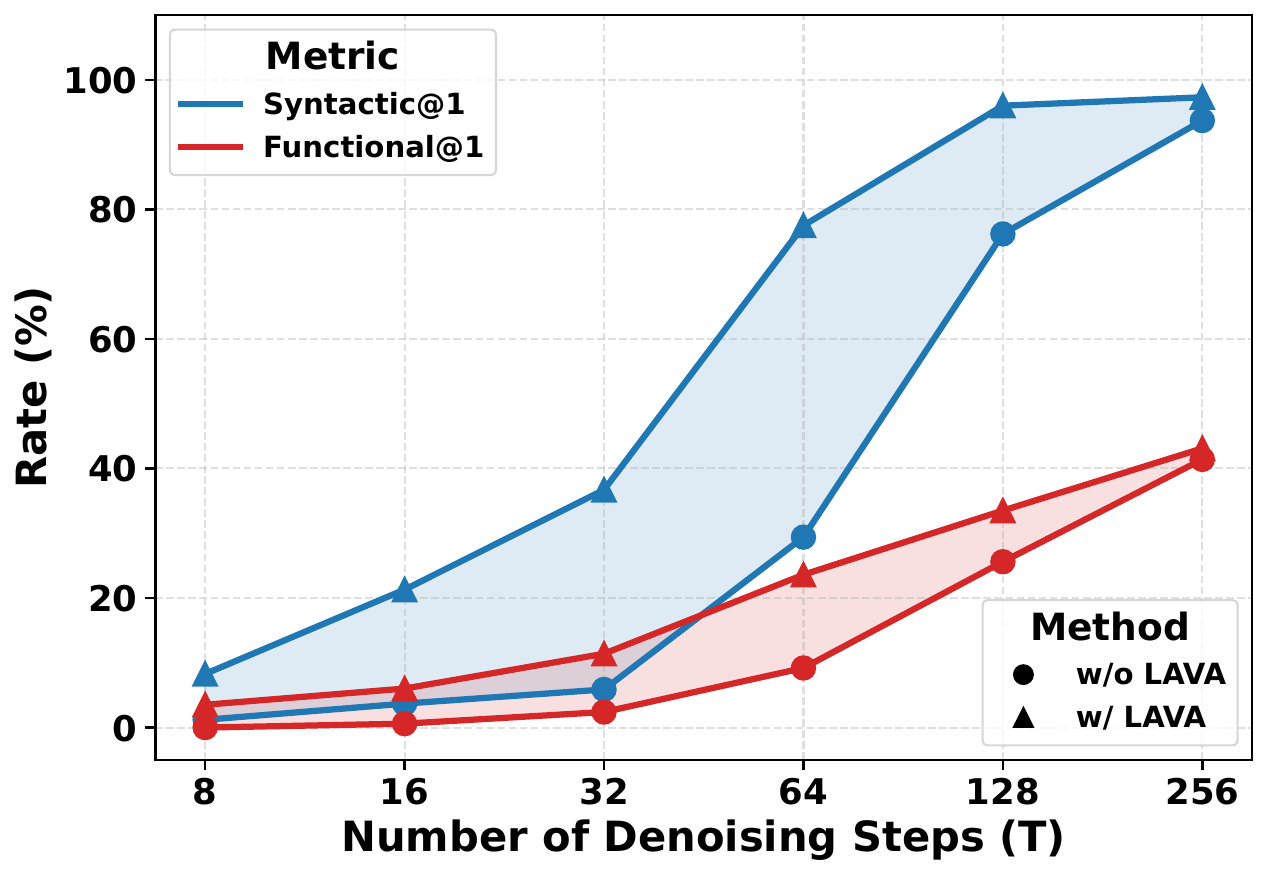}%
        \put(84.8,15.0){\begingroup\setlength{\fboxsep}{0pt}\colorbox{white}{\makebox[31pt][l]{\rule[-1.2pt]{0pt}{7.2pt}\fontsize{5.5pt}{6pt}\selectfont\sffamily\bfseries w/o LAVE}}\endgroup}
        \put(84.8,11.2){\begingroup\setlength{\fboxsep}{0pt}\colorbox{white}{\makebox[31pt][l]{\rule[-1.2pt]{0pt}{7.2pt}\fontsize{5.5pt}{6pt}\selectfont\sffamily\bfseries w/ LAVE}}\endgroup}
    \end{overpic}%
    \vspace{-0.03in}
    \caption{\cc
    \dataname{CPP-Bench}}
    \label{fig:trade-off-cpp}
\end{subfigure}
\vspace{-0.11in}
\caption{\cc
Performance of \name under different numbers of denoising steps $T$ on \dataname{JSON-Bench} and \dataname{CPP-Bench} using \llmname{Dream-7B}.}
\label{fig:trade-off}
\vspace{-0.14in}
\end{figure}

Although \name is primarily designed to improve the syntactic correctness of dLLM outputs, its effect can also be understood from a correctness--efficiency perspective. The efficiency advantage of dLLMs mainly comes from parallel decoding: with fewer denoising steps, the model can generate more tokens per step. However, this efficiency often comes at the cost of lower correctness. Therefore, dLLM decoding naturally involves a trade-off between correctness and efficiency.

\name improves this trade-off by increasing correctness under the same denoising budget. To examine this effect, we evaluate \name with \llmname{Dream-7B} on \dataname{JSON-Bench} and \dataname{CPP-Bench} under different numbers of denoising steps, as shown in Figure~\ref{fig:trade-off}. Compared with unconstrained decoding, \name achieves higher correctness across different denoising budgets.
Therefore, \name may not be viewed only as a method that improves syntactic correctness. It also provides a better correctness--efficiency trade-off, making dLLMs more practical in latency-sensitive scenarios.

\end{cp}

\subsection{Threats to Validity}
\label{sec:threats}

We identify three primary threats to the validity of our study:

\ding{182} \textbf{Generalizability of the Experimental Results.}
A potential concern is whether our findings hold across diverse tasks and models. To mitigate this threat, we carefully selected a broad range of benchmarks and models for our evaluation. Specifically, we adopted \cl{five} representative benchmarks covering code generation, information extraction, and scientific applications. These domains reflect typical real-world scenarios where structured outputs are essential, aligning with setups in prior research~\cite{igcd}. Furthermore, we evaluated our approach on four widely used dLLMs, ensuring that our conclusions are not artifacts of a specific model architecture or training data distribution. 

\ding{183} \textbf{Reliability-Oriented Constraint Design.}
Due to the non-autoregressive generation mechanism of dLLMs, it is inherently difficult to guarantee that all accepted tokens are valid while all rejected tokens are invalid. As a result, our approach may occasionally reject proposed tokens that could still lead to valid completions. Nevertheless, we argue that the primary goal of constrained decoding is to improve the reliability of generated outputs. From this perspective, ensuring that all accepted tokens preserve grammatical validity is more important than accepting every potentially valid token.
Empirically, Section~\ref{sec:pre} suggests that a small lookahead budget is usually sufficient to uncover an extendable candidate when one exists, reducing the likelihood of rejecting valid tokens. The results in Section~\ref{sec:rq1} and Section~\ref{sec:rq2} further show that our method consistently improves syntactic correctness and also increases functional correctness, indicating that the practical impact of rejecting some valid tokens is negligible.

\ding{184} \textbf{Impact of Hyperparameters.}
Our approach introduces two hyperparameters, which could potentially limit utility if the approach requires extensive tuning. To address this, we initially selected reasonable values based on intuition and observed consistent performance improvements across multiple benchmarks and models. To rigorously examine the sensitivity of these parameters, we conducted a sensitivity analysis by varying their values over a wide range. The empirical results indicate that our approach demonstrates strong robustness to variations in hyperparameter settings, maintaining high performance without the need for precise tuning.


\vspace{-4pt}
\section{Conclusion}
In this paper, we present \name, a new constrained decoding approach tailored to emerging diffusion LLMs.
\name exploits the unique ability of these models to predict token distributions for all positions in parallel during a single forward pass.
Our approach addresses the limitations of existing constrained decoding methods by enforcing reliable grammar constraints that preserve the potential for intermediate outputs to be extended into valid sentences. Experimental results on code generation, JSON generation, and chemical expression generation demonstrate that \name can effectively improve the reliability of dLLMs in producing CFG-compliant outputs.

\vspace{-4pt}
\section*{Data Availability}

The source code is provided at: \url{https://github.com/THU-Agent/LAVE}.

\section*{Acknowledgments}

This work was supported by the Tsinghua University--Keystone Electrical (Zhejiang) Co., Ltd. Joint Research Center for Embodied Multimodal Artificial Intelligence (JCEMAI) and the Beijing Natural Science Foundation under Grant No. 4264107. 

This work was also supported by the Beijing Natural Science Foundation under Grant No. QY24136, the National Natural Science Foundation of China under Grant No. 62192733, 62192730, 62192731, the National Key R\&D Program under Grant No. 2023YFB4503801, the Beijing Major Science and Technology Project under Contract no. Z251100008425005.

\bibliographystyle{ACM-Reference-Format}
\bibliography{ref}

@article{mercury,
  author       = {Samar Khanna and
                  Siddhant Kharbanda and
                  Shufan Li and
                  Harshit Varma and
                  Eric Wang and
                  Sawyer Birnbaum and
                  Ziyang Luo and
                  Yanis Miraoui and
                  Akash Palrecha and
                  Stefano Ermon and
                  Aditya Grover and
                  Volodymyr Kuleshov},
  title        = {Mercury: Ultra-Fast Language Models Based on Diffusion},
  journal      = {CoRR},
  volume       = {abs/2506.17298},
  year         = {2025},
  url          = {https://doi.org/10.48550/arXiv.2506.17298},
  doi          = {10.48550/ARXIV.2506.17298},
  eprinttype   = {arXiv},
  eprint       = {2506.17298},
  bibsource    = {dblp computer science bibliography, https://dblp.org}
}

@article{llada,
  title={Large language diffusion models},
  author={Nie, Shen and Zhu, Fengqi and You, Zebin and Zhang, Xiaolu and Ou, Jingyang and Hu, Jun and Zhou, Jun and Lin, Yankai and Wen, Ji-Rong and Li, Chongxuan},
  journal={arXiv preprint arXiv:2502.09992},
  year={2025}
}

@article{llada15,
  title={LLaDA 1.5: Variance-Reduced Preference Optimization for Large Language Diffusion Models},
  author={Zhu, Fengqi and Wang, Rongzhen and Nie, Shen and Zhang, Xiaolu and Wu, Chunwei and Hu, Jun and Zhou, Jun and Chen, Jianfei and Lin, Yankai and Wen, Ji-Rong and others},
  journal={arXiv preprint arXiv:2505.19223},
  year={2025}
}

@article{dream,
  title={Dream 7b: Diffusion large language models},
  author={Ye, Jiacheng and Xie, Zhihui and Zheng, Lin and Gao, Jiahui and Wu, Zirui and Jiang, Xin and Li, Zhenguo and Kong, Lingpeng},
  journal={arXiv preprint arXiv:2508.15487},
  year={2025}
}

@article{dreamcoder,
  title={Dream-coder 7b: An open diffusion language model for code},
  author={Xie, Zhihui and Ye, Jiacheng and Zheng, Lin and Gao, Jiahui and Dong, Jingwei and Wu, Zirui and Zhao, Xueliang and Gong, Shansan and Jiang, Xin and Li, Zhenguo and others},
  journal={arXiv preprint arXiv:2509.01142},
  year={2025}
}

@article{diffucoder,
  title={DiffuCoder: Understanding and Improving Masked Diffusion Models for Code Generation},
  author={Gong, Shansan and Zhang, Ruixiang and Zheng, Huangjie and Gu, Jiatao and Jaitly, Navdeep and Kong, Lingpeng and Zhang, Yizhe},
  journal={arXiv preprint arXiv:2506.20639},
  year={2025}
}

@article{zhang2025beyond,
  title={Beyond Autoregression: An Empirical Study of Diffusion Large Language Models for Code Generation},
  author={Li, Chengze and Zhang, Yitong and Li, Jia and Cai, Liyi and Li, Ge and others},
  journal={arXiv preprint arXiv:2509.11252},
  year={2025}
}

@article{yang2025difftester,
  title={DiffTester: Accelerating Unit Test Generation for Diffusion LLMs via Repetitive Pattern},
  author={Yang, Lekang and Liu, Yuetong and Zhang, Yitong and Li, Jia},
  journal={arXiv preprint arXiv:2509.24975},
  year={2025}
}

@article{igcd,
  title={Constrained decoding of diffusion llms with context-free grammars},
  author={M{\"u}ndler, Niels and Dekoninck, Jasper and Vechev, Martin},
  journal={arXiv preprint arXiv:2508.10111},
  year={2025}
}

@inproceedings{humaneval-x,
  title={Codegeex: A pre-trained model for code generation with multilingual benchmarking on humaneval-x},
  author={Zheng, Qinkai and Xia, Xiao and Zou, Xu and Dong, Yuxiao and Wang, Shan and Xue, Yufei and Shen, Lei and Wang, Zihan and Wang, Andi and Li, Yang and others},
  booktitle={Proceedings of the 29th ACM SIGKDD Conference on Knowledge Discovery and Data Mining},
  pages={5673--5684},
  year={2023}
}

@article{willard2023efficient,
  title={Efficient guided generation for large language models},
  author={Willard, Brandon T and Louf, R{\'e}mi},
  journal={arXiv preprint arXiv:2307.09702},
  year={2023}
}

@article{chen2025pre,
  title={Pre$^{3}$: Enabling Deterministic Pushdown Automata for Faster Structured LLM Generation},
  author={Chen, Junyi and Bai, Shihao and Wang, Zaijun and Wu, Siyu and Du, Chuheng and Yang, Hailong and Gong, Ruihao and Liu, Shengzhong and Wu, Fan and Chen, Guihai},
  journal={arXiv preprint arXiv:2506.03887},
  year={2025}
}

@article{beurer2024guiding,
  title={Guiding llms the right way: Fast, non-invasive constrained generation},
  author={Beurer-Kellner, Luca and Fischer, Marc and Vechev, Martin},
  journal={arXiv preprint arXiv:2403.06988},
  year={2024}
}

@article{ugare2024syncode,
  title={SynCode: LLM generation with grammar augmentation},
  author={Ugare, Shubham and Suresh, Tarun and Kang, Hangoo and Misailovic, Sasa and Singh, Gagandeep},
  journal={Transactions on Machine Learning Research},
  year={2024}
}

@misc{llguidance,
  title        = {LLGuidance},
  author       = {Microsoft},
  year         = {2025},
  month        = jun,
  url = {https://github.com/guidance-ai/llguidance?tab=readme-ov-file},
}

@article{sun2025earley,
  title={Earley-Driven Dynamic Pruning for Efficient Structured Decoding},
  author={Sun, Xintong and Wei, Chi and Tian, Minghao and Ni, Shiwen},
  journal={arXiv preprint arXiv:2506.01151},
  year={2025}
}

@article{gong2024scaling,
  title={Scaling diffusion language models via adaptation from autoregressive models},
  author={Gong, Shansan and Agarwal, Shivam and Zhang, Yizhe and Ye, Jiacheng and Zheng, Lin and Li, Mukai and An, Chenxin and Zhao, Peilin and Bi, Wei and Han, Jiawei and others},
  journal={arXiv preprint arXiv:2410.17891},
  year={2024}
}

@article{gao2025self,
  title={Self Speculative Decoding for Diffusion Large Language Models},
  author={Gao, Yifeng and Ji, Ziang and Wang, Yuxuan and Qi, Biqing and Xu, Hanlin and Zhang, Linfeng},
  journal={arXiv preprint arXiv:2510.04147},
  year={2025}
}

@article{li2025survey,
  title={A survey on diffusion language models},
  author={Li, Tianyi and Chen, Mingda and Guo, Bowei and Shen, Zhiqiang},
  journal={arXiv preprint arXiv:2508.10875},
  year={2025}
}

@article{arriola2025block,
  title={Block diffusion: Interpolating between autoregressive and diffusion language models},
  author={Arriola, Marianne and Gokaslan, Aaron and Chiu, Justin T and Yang, Zhihan and Qi, Zhixuan and Han, Jiaqi and Sahoo, Subham Sekhar and Kuleshov, Volodymyr},
  journal={arXiv preprint arXiv:2503.09573},
  year={2025}
}

@article{yang2025mmada,
  title={Mmada: Multimodal large diffusion language models},
  author={Yang, Ling and Tian, Ye and Li, Bowen and Zhang, Xinchen and Shen, Ke and Tong, Yunhai and Wang, Mengdi},
  journal={arXiv preprint arXiv:2505.15809},
  year={2025}
}

@article{d2f,
  title={Diffusion llms can do faster-than-ar inference via discrete diffusion forcing},
  author={Wang, Xu and Xu, Chenkai and Jin, Yijie and Jin, Jiachun and Zhang, Hao and Deng, Zhijie},
  journal={arXiv preprint arXiv:2508.09192},
  year={2025}
}

@article{hui2024qwen2,
  title={Qwen2. 5-coder technical report},
  author={Hui, Binyuan and Yang, Jian and Cui, Zeyu and Yang, Jiaxi and Liu, Dayiheng and Zhang, Lei and Liu, Tianyu and Zhang, Jiajun and Yu, Bowen and Lu, Keming and others},
  journal={arXiv preprint arXiv:2409.12186},
  year={2024}
}

@article{touvron2023llama,
  title={Llama: Open and efficient foundation language models},
  author={Touvron, Hugo and Lavril, Thibaut and Izacard, Gautier and Martinet, Xavier and Lachaux, Marie-Anne and Lacroix, Timoth{\'e}e and Rozi{\`e}re, Baptiste and Goyal, Naman and Hambro, Eric and Azhar, Faisal and others},
  journal={arXiv preprint arXiv:2302.13971},
  year={2023}
}

@article{earley,
  title={An efficient context-free parsing algorithm},
  author={Earley, Jay},
  journal={Communications of the ACM},
  volume={13},
  number={2},
  pages={94--102},
  year={1970},
  publisher={ACM New York, NY, USA}
}

@article{deremer1971simple,
  title={Simple LR (k) grammars},
  author={DeRemer, Franklin L},
  journal={Communications of the ACM},
  volume={14},
  number={7},
  pages={453--460},
  year={1971},
  publisher={ACM New York, NY, USA}
}

@article{wang2024fantastic,
  title={FANTAstic SEquences and where to find them: Faithful and efficient API call generation through state-tracked constrained decoding and reranking},
  author={Wang, Zhuoer and Ribeiro, Leonardo FR and Papangelis, Alexandros and Mukherjee, Rohan and Wang, Tzu-Yen and Zhao, Xinyan and Biswas, Arijit and Caverlee, James and Metallinou, Angeliki},
  journal={arXiv preprint arXiv:2407.13945},
  year={2024}
}

@article{tracerl,
  title={Revolutionizing reinforcement learning framework for diffusion large language models},
  author={Wang, Yinjie and Yang, Ling and Li, Bowen and Tian, Ye and Shen, Ke and Wang, Mengdi},
  journal={arXiv preprint arXiv:2509.06949},
  year={2025}
}

@article{geng2023grammar,
  title={Grammar-constrained decoding for structured NLP tasks without finetuning},
  author={Geng, Saibo and Josifoski, Martin and Peyrard, Maxime and West, Robert},
  journal={arXiv preprint arXiv:2305.13971},
  year={2023}
}

@article{loula2025syntactic,
  title={Syntactic and semantic control of large language models via sequential monte carlo},
  author={Loula, Jo{\~a}o and LeBrun, Benjamin and Du, Li and Lipkin, Ben and Pasti, Clemente and Grand, Gabriel and Liu, Tianyu and Emara, Yahya and Freedman, Marjorie and Eisner, Jason and others},
  journal={arXiv preprint arXiv:2504.13139},
  year={2025}
}

@article{aho1974lr,
  title={LR parsing},
  author={Aho, Alfred V. and Johnson, Stephen C.},
  journal={ACM Computing Surveys (CSUR)},
  volume={6},
  number={2},
  pages={99--124},
  year={1974},
  publisher={ACM New York, NY, USA}
}

@article{wagner1998efficient,
  title={Efficient and flexible incremental parsing},
  author={Wagner, Tim A and Graham, Susan L},
  journal={ACM Transactions on Programming Languages and Systems (TOPLAS)},
  volume={20},
  number={5},
  pages={980--1013},
  year={1998},
  publisher={ACM New York, NY, USA}
}

@article{parys2025constrained,
  title={Constrained Adaptive Rejection Sampling},
  author={Parys, Pawe{\'L} and Vaidya, Sairam and Berg-Kirkpatrick, Taylor and D'Antoni, Loris},
  journal={arXiv preprint arXiv:2510.01902},
  year={2025}
}

@article{jiang2024rocode,
  title={Rocode: Integrating backtracking mechanism and program analysis in large language models for code generation},
  author={Jiang, Xue and Dong, Yihong and Tao, Yongding and Liu, Huanyu and Jin, Zhi and Jiao, Wenpin and Li, Ge},
  journal={arXiv preprint arXiv:2411.07112},
  year={2024}
}

@article{ugare2024itergen,
  title={IterGen: Iterative Semantic-aware Structured LLM Generation with Backtracking},
  author={Ugare, Shubham and Gumaste, Rohan and Suresh, Tarun and Singh, Gagandeep and Misailovic, Sasa},
  journal={arXiv preprint arXiv:2410.07295},
  year={2024}
}

@article{melcer2024constrained,
  title={Constrained decoding for code language models via efficient left and right quotienting of context-sensitive grammars},
  author={Melcer, Daniel and Fulton, Nathan and Gouda, Sanjay Krishna and Qian, Haifeng},
  journal={CoRR},
  year={2024}
}

@article{park2025flexible,
  title={Flexible and efficient grammar-constrained decoding},
  author={Park, Kanghee and Zhou, Timothy and D'Antoni, Loris},
  journal={arXiv preprint arXiv:2502.05111},
  year={2025}
}

@article{humaneval,
  title={Evaluating large language models trained on code},
  author={Chen, Mark},
  journal={arXiv preprint arXiv:2107.03374},
  year={2021}
}

@misc{nous_json_mode_eval,
  author       = {{NousResearch}},
  title        = {json-mode-eval},
  howpublished = {\url{https://huggingface.co/datasets/NousResearch/json-mode-eval}},
  year         = {2024},
  note         = {Hugging Face Datasets}
}

@article{nguyen2025attention,
  title={Attention is all you need for kv cache in diffusion llms},
  author={Nguyen-Tri, Quan and Ranjan, Mukul and Shen, Zhiqiang},
  journal={arXiv preprint arXiv:2510.14973},
  year={2025}
}

@article{hu2025accelerating,
  title={Accelerating diffusion language model inference via efficient kv caching and guided diffusion},
  author={Hu, Zhanqiu and Meng, Jian and Akhauri, Yash and Abdelfattah, Mohamed S and Seo, Jae-sun and Zhang, Zhiru and Gupta, Udit},
  journal={arXiv preprint arXiv:2505.21467},
  year={2025}
}

@article{sahoo2024simple,
  title={Simple and effective masked diffusion language models},
  author={Sahoo, Subham and Arriola, Marianne and Schiff, Yair and Gokaslan, Aaron and Marroquin, Edgar and Chiu, Justin and Rush, Alexander and Kuleshov, Volodymyr},
  journal={Advances in Neural Information Processing Systems},
  volume={37},
  pages={130136--130184},
  year={2024}
}

@misc{gemini_diffusion,
      title={Gemini Diffusion}, 
      author={Google DeepMind},
      year={2025},
      url={https://deepmind.google/models/gemini-diffusion/}, 
}

@article{ugare2024improving,
  title={Improving llm code generation with grammar augmentation},
  author={Ugare, Shubham and Suresh, Tarun and Kang, Hangoo and Misailovic, Sasa and Singh, Gagandeep},
  journal={arXiv preprint arXiv:2403.01632},
  volume={19},
  year={2024}
}

@article{suresh2025dingo,
  title={DINGO: Constrained Inference for Diffusion LLMs},
  author={Suresh, Tarun and Banerjee, Debangshu and Ugare, Shubham and Misailovic, Sasa and Singh, Gagandeep},
  journal={arXiv preprint arXiv:2505.23061},
  year={2025}
}

@inproceedings{dong2025rethinking,
  title={Rethinking repetition problems of llms in code generation},
  author={Dong, Yihong and Liu, Yuchen and Jiang, Xue and Gu, Bin and Jin, Zhi and Li, Ge},
  booktitle={Proceedings of the 63rd Annual Meeting of the Association for Computational Linguistics (Volume 1: Long Papers)},
  pages={965--985},
  year={2025}
}

@inproceedings{fu2021theoretical,
  title={A theoretical analysis of the repetition problem in text generation},
  author={Fu, Zihao and Lam, Wai and So, Anthony Man-Cho and Shi, Bei},
  booktitle={Proceedings of the AAAI Conference on Artificial Intelligence},
  volume={35},
  number={14},
  pages={12848--12856},
  year={2021}
}

@article{holtzman2019curious,
  title={The curious case of neural text degeneration},
  author={Holtzman, Ari and Buys, Jan and Du, Li and Forbes, Maxwell and Choi, Yejin},
  journal={arXiv preprint arXiv:1904.09751},
  year={2019}
}

@article{li2023repetition,
  title={Repetition in repetition out: Towards understanding neural text degeneration from the data perspective},
  author={Li, Huayang and Lan, Tian and Fu, Zihao and Cai, Deng and Liu, Lemao and Collier, Nigel and Watanabe, Taro and Su, Yixuan},
  journal={Advances in Neural Information Processing Systems},
  volume={36},
  pages={72888--72903},
  year={2023}
}

@article{xu2022learning,
  title={Learning to break the loop: Analyzing and mitigating repetitions for neural text generation},
  author={Xu, Jin and Liu, Xiaojiang and Yan, Jianhao and Cai, Deng and Li, Huayang and Li, Jian},
  journal={Advances in Neural Information Processing Systems},
  volume={35},
  pages={3082--3095},
  year={2022}
}

@article{li2025diffuguard,
  title={Diffuguard: How intrinsic safety is lost and found in diffusion large language models},
  author={Li, Zherui and Nie, Zheng and Zhou, Zhenhong and Guo, Yufei and Liu, Yue and Zhang, Yitong and Cheng, Yu and Wen, Qingsong and Wang, Kun and Zhang, Jiaheng},
  journal={arXiv preprint arXiv:2509.24296},
  year={2025}
}

@article{cheng2025deer,
  title={DEER: Draft with Diffusion, Verify with Autoregressive Models},
  author={Cheng, Zicong and Yang, Guo-Wei and Li, Jia and Deng, Zhijie and Guo, Meng-Hao and Hu, Shi-Min},
  journal={arXiv preprint arXiv:2512.15176},
  year={2025}
}

@article{liu2025wedlm,
  title={WeDLM: Reconciling Diffusion Language Models with Standard Causal Attention for Fast Inference},
  author={Liu, Aiwei and He, Minghua and Zeng, Shaoxun and Zhang, Sijun and Zhang, Linhao and Wu, Chuhan and Jia, Wei and Liu, Yuan and Zhou, Xiao and Zhou, Jie},
  journal={arXiv preprint arXiv:2512.22737},
  year={2025}
}

@article{song2025seed,
  title={Seed diffusion: A large-scale diffusion language model with high-speed inference},
  author={Song, Yuxuan and Zhang, Zheng and Luo, Cheng and Gao, Pengyang and Xia, Fan and Luo, Hao and Li, Zheng and Yang, Yuehang and Yu, Hongli and Qu, Xingwei and others},
  journal={arXiv preprint arXiv:2508.02193},
  year={2025}
}

@inproceedings{nakshatri2025constrained,
  title={Constrained decoding with speculative lookaheads},
  author={Nakshatri, Nishanth Sridhar and Roy, Shamik and Das, Rajarshi and Chaidaroon, Suthee and Boytsov, Leonid and Gangadharaiah, Rashmi},
  booktitle={Proceedings of the 2025 Conference of the Nations of the Americas Chapter of the Association for Computational Linguistics: Human Language Technologies (Volume 1: Long Papers)},
  pages={4681--4700},
  year={2025}
}

@article{zhu2026dllm,
  title={ES-dLLM: Efficient Inference for Diffusion Large Language Models by Early-Skipping},
  author={Zhu, Zijian and Ren, Fei and Tan, Zhanhong and Ma, Kaisheng},
  journal={arXiv preprint arXiv:2603.10088},
  year={2026}
}

@article{zhang2026grammar,
  title={Grammar-Constrained Decoding Can Jailbreak LLMs into Generating Malicious Code},
  author={Zhang, Yitong and Lu, Shiteng and Li, Jia},
  journal={arXiv preprint arXiv:2606.11817},
  year={2026}
}

\end{document}